\documentclass[10pt]{article} 
\usepackage[preprint]{tmlr}

\usepackage{amsmath,amsfonts,bm}

\def\eqref#1{equation~\ref{#1}}

\def\1{\bm{1}}

\DeclareMathAlphabet{\mathsfit}{\encodingdefault}{\sfdefault}{m}{sl}
\SetMathAlphabet{\mathsfit}{bold}{\encodingdefault}{\sfdefault}{bx}{n}

\usepackage{hyperref}
\usepackage{url}

\usepackage{graphicx}
\usepackage[linesnumbered,ruled,vlined]{algorithm2e}
\usepackage{amsmath}
\usepackage{cleveref}
\usepackage{makecell}
\usepackage{multirow}
\usepackage{subcaption}
\usepackage{multicol}
\usepackage{graphicx} 
\usepackage{tikz}
\usepackage{listofitems} 
\usepackage{xcolor}
\usepackage{makecell}
\usetikzlibrary{shapes.callouts}
\usepackage{tabularx}
\usetikzlibrary{shapes.geometric} 
\usetikzlibrary{positioning,fit}
\usetikzlibrary{calc} 
\usetikzlibrary{matrix, positioning, calc}
\usepackage[normalem]{ulem} 

\newcommand\restr[2]{{
  \left.\kern-\nulldelimiterspace 
  #1 
  \littletaller 
  \right|_{#2} 
  }}
\newcommand{\littletaller}{\mathchoice{\vphantom{\big|}}{}{}{}}

\newcommand{\messagebubble}[1]{
    \begin{tikzpicture}[baseline=(bubble.base)]
    \node[draw, rectangle, rounded corners, fill=gray!20, align=left, minimum width=1cm, minimum height=0.8cm, inner sep=4pt] (bubble) {#1};
    \end{tikzpicture}
}

\title{Detecting Systematic Weaknesses in Vision Models along Predefined Human-Understandable Dimensions}

\author{\name Sujan Sai Gannamaneni \email sujan.sai.gannamaneni@iais.fraunhofer.de \\
      Fraunhofer IAIS, Lamarr Institute
      \AND
      \name Rohil Prakash Rao \email rohil.prakash.rao@iais.fraunhofer.de \\
      Fraunhofer IAIS
      \AND
      \name Michael Mock \email michael.mock@iais.fraunhofer.de\\
      Fraunhofer IAIS
      \AND
      \name Maram Akila \email maram.akila@iais.fraunhofer.de\\
      Fraunhofer IAIS, Lamarr Institute
      \AND
      \name Stefan Wrobel \email stefan.wrobel@iais.fraunhofer.de\\
      Fraunhofer IAIS, University of Bonn
      }

\def\month{MM}  
\def\year{YYYY} 
\def\openreview{\url{https://openreview.net/forum?id=XXXX}} 

\begin{document}

\maketitle

\begin{abstract}
Slice discovery methods (SDMs) are prominent algorithms for finding systematic weaknesses in DNNs. They identify top-k semantically coherent slices/subsets of data where a DNN-under-test has low performance. For being directly useful, slices should be aligned with human-understandable and relevant dimensions, which, for example, are defined by safety and domain experts as part of the operational design domain (ODD). While SDMs can be applied effectively on structured data, their application on image data is complicated by the lack of semantic metadata. To address these issues, we present an algorithm that combines foundation models for zero-shot image classification to generate semantic metadata with methods for combinatorial search to find systematic weaknesses in images. In contrast to existing approaches, ours identifies weak slices that are in line with pre-defined human-understandable dimensions. As the algorithm includes foundation models, its intermediate and final results may not always be exact. Therefore, we include an approach to address the impact of noisy metadata. We validate our algorithm on both synthetic and real-world datasets, demonstrating its ability to recover human-understandable systematic weaknesses. Furthermore, using our approach, we identify systematic weaknesses of multiple pre-trained and publicly available state-of-the-art computer vision DNNs.

\end{abstract}

\section{Introduction}
\label{sec:intro}

With recent advances in machine learning (ML), there has been a significant improvement in the modeling of unstructured data, such as images. 
However, for safety-critical applications, ML models need to be developed with a focus on trustworthiness by investigating and correcting potential failure modes.
To that end, systematic errors of DNNs need to be studied and rectified. Hidden stratification~\citep{oakden2020hidden} and fairness-related bias~\citep{buolamwini2018gender, wang2020towards, li2023dark} due to spurious correlations~\citep{xiao2020noise, geirhos2020shortcut,mahmood2021detecting} and underrepresented subpopulations~\citep{santurkar2020breeds,sagawa2019distributionally} are some examples of potential failure modes where the error or weakness is systematic in nature. The existence of these modes implies that there are slices~\footnote{In the literature, slices are often also called subgroups or subsets of data. All three terms are used interchangeably in relation to systematic weakness analysis.} of data where the performance of the DNN-under-test (\textbf{DuT}) is worse than the average performance on the entire test dataset. Although identifying slices with weak performance would be trivial by simply grouping samples on which models have high error, identifying slices that are both semantically coherent and have high error is challenging. This is due to the lack of semantic metadata that describes the slices for many data domains (e.g., images, text). Despite this challenge, identifying such slices provides a human-understandable global explanation of the model behavior. Moreover, semantically coherent weak slices offer actionable insights for debugging and auditing models. 


From a safety and certification perspective, upcoming standards (e.g., ISO/PAS 8800~\citep{ISO8800:2024}), and works with a focus on AI in automotive~\citep{Koopman2019HowMO, burton2022safety}, aerospace~\citep{EASA_concept_paper} and railway~\citep{zeller2023safety} domains have highlighted the importance of data completeness and quality using, in most cases, Operational Design Domains (ODDs). 
In automotive,~\citet{herrmann2022using} have proposed ontologies for different traffic participants that can be used to build ODDs for automated driving.  
The goal of using such ODDs is to describe the scope of AI applications in terms of human-understandable, safety-relevant dimensions where comprehensible safety argumentations can be built w.r.t.\ robustness, explainability, and interpretability.  
To facilitate building such safety augmentations, testing approaches for ML developers and safety experts that evaluate DNN performance and identify systematic weaknesses are essential.

Although in recent years, several works~\citep{chung2019slice, sagadeeva2021sliceline, d2022spotlight, eyuboglu2022domino, Metzen_2023_ICCV, plumb2023towards, jain2023distilling, gao2023adaptive, NEURIPS2023_0f53ecc0} have proposed methods for analyzing systematic weaknesses, there is a lack of focus on identifying weaknesses of models evaluated on real-world datasets, where the weaknesses align with human-understandable semantic concepts defined, for example, by safety experts in ODDs.
We argue that it is more beneficial from a safety perspective if the approaches to identify systematic weaknesses are ODD-compliant for two main reasons: (i) the slices are \textbf{useful} as the identified vulnerabilities are aligned with human-understandable safety-relevant dimensions, and (ii) the slices are \textbf{actionable} as ML developers can gather more data to retrain or reweight existing samples to improve performance along the safety-relevant dimensions. We address the challenge of analyzing unstructured image data by designing an algorithm that leverages recent advances in foundational models and systematic weakness analysis methods for structured data. Our contributions can be summarized as follows:
\begin{itemize}
    \item We introduce an algorithm that takes in an image dataset, ODD description and performance values of a \textbf{DuT} as inputs and outputs systematic weaknesses of the \textbf{DuT} (see~\cref{sec:method}).    
    \item Concretely, as part of the metadata generation module, we make use of CLIP~\citep{radford2021learning} to leverage its rich joint image, text embedding space. As part of the slice discovery module, we propose using SliceLine~\citep{sagadeeva2021sliceline} with modifications to identify weak slices that align with the ODD (see~\cref{sec:method}).
    \item In addition, we address the noisy nature of metadata generation and propose a way to recover relevant weak slices even if CLIP labeling is suboptimal. We empirically evaluate the behavior of our algorithm at various levels of label quality using synthetic data (see~\cref{sec:synthetic_data}). 
    \item Furthermore, we evaluate multiple pre-trained and publicly available DNNs-under-test using our algorithm on real-world datasets and provide insights into their systematic weaknesses (see~\cref{sec:results}). 
\end{itemize}


\section{Related Work}
\label{sec:related_work}

In this section, we review the recent progress in analyzing systematic weaknesses using slice discovery methods (SDMs)~\citep{eyuboglu2022domino} for structured and unstructured data and highlight their connection to interpretability and feature attribution methods.  

For structured data, methods such as SliceFinder~\citep{chung2019slice} and SliceLine~\citep{sagadeeva2021sliceline} leverage the rich metadata available in the form of features to slice the data and exhaustively search for top-k low-performing slices. The differences between these two approaches lie in the scoring of errors, the pruning strategy, and how they handle slice sizes. 
Although these two approaches were explicitly developed to identify systematic weaknesses, subgroup-discovery techniques~\citep{atzmueller2015subgroup}, a subset of data mining, have a similar problem formulation and could also potentially be used for slice discovery of structured data. 

For unstructured data such as images, where metadata is not directly available, SOTA approaches have taken two lines of research. In the first line of prior work, for a given test dataset, DNN embeddings are used as proxies for coherency. Weak-performing slices of the data are obtained by clustering these embeddings along with model errors. 
Here, approaches such as Spotlight~\citep{d2022spotlight} perform clustering on the embeddings of the final layers of the \textbf{DuT} itself. In contrast, recent approaches leverage the joint embedding space of foundational models such as CLIP~\citep{radford2021learning} and apply mixture models like in DOMINO~\citep{eyuboglu2022domino} or SVMs like in SVM-FD~\citep{jain2023distilling} to identify coherent clusters. In Spotlight, an additional step involving humans is required to inspect and understand what uniquely constitutes a weak slice. 
DOMINO and SVM-FD automate the slice description process to reduce human effort and bias using an additional DNN. 
In all these approaches, as coherence is only loosely enforced based on DNN embeddings, it is not always clear what specific human-understandable concept uniquely constitutes a slice. Without this knowledge, it would be unclear to the ML developers what new data samples would need to be collected to retrain the model and fix the systematic weakness. 
To mitigate this problem, some approaches~\citep{gao2023adaptive, slyman2023vlslice} propose iterative human-in-the-loop testing to ensure that the identified slices are human-understandable. 
 
In the second line of prior work, inspired by counterfactuals and leveraging CLIP, several approaches~\citep{wiles2022discovering, Metzen_2023_ICCV} propose synthetically generating new (counterfactual) images that would lead to erroneous predictions by controlling the content and data shift in the image.
Among these, PromptAttack~\citep{Metzen_2023_ICCV} also proposes to identify weaknesses that are aligned with the ODDs. However, while PromptAttack generates new samples using image-generation DNNs, which could potentially introduce biases due to domain shift, our approach is more closely aligned with earlier methods that evaluate a DNN on a given test dataset.
In this direction, HiBug~\citep{NEURIPS2023_0f53ecc0} utilizes a GPT-based model to assign attributes to a given dataset. Building on this and appearing concurrently with our work, DebugAgent~\citep{chen2025debugagent} extends HiBug with a search algorithm to identify weak slices. While we also apply attributes to the data to perform a subsequent weak slice search, we, instead, opt for the less compute-intensive CLIP~\cite{radford2021learning} model to generate attributes. Additionally, we develop a Bayesian framework to compensate for the label noise that occurs from the attribution.

In contrast to SDMs, local interpretability and feature attribution methods~\citep{ribeiro2016should, lundberg2017unified}, while linking achieved understandability to actionability~\citep{guidotti2022stable}, identify local explanations and not the global systematic weaknesses. In addition, the feature attribution methods themselves might not always be robust or consistent~\citep{krishna2022disagreement}.

\section{Method}
\label{sec:method}

\begin{figure*}[!bht] 
  \centering

    \input{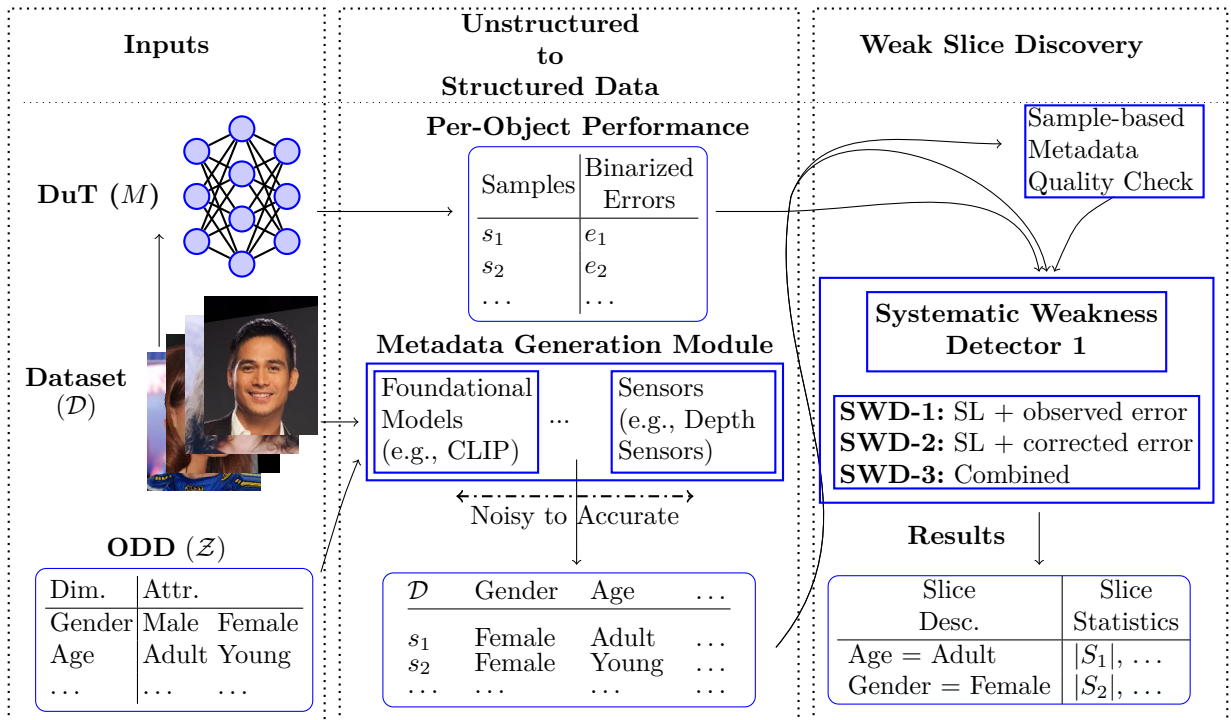}
  \caption{Our algorithm for finding systematic weaknesses of CV models. Given a model, a test dataset, and an ODD description for the objects we are interested in, we build a database of object-level performance and metadata in a structured format. Weak slice discovery methods are then applied to this database to identify top-k weak slices of the model.}
  \label{fig:workflow}
\end{figure*}

In this section, we present our algorithm for weakness detection on the basis of human-understandable semantic dimensions. To this end, we introduce notation regarding metadata and slicing, discuss the generation of metadata, formulate DNN weakness within a Bayesian framework to account for the impact of noise, and lastly detail how such impact can be acknowledged within slice discovery algorithms.


\textbf{Notation}: Consider a DNN-under-test (\textbf{DuT}) $M$ trained on some computer vision task. Let $\mathcal D$ be the (test) dataset containing the inputs and the corresponding task-related ground truth.
For each sample $s_i\in\mathcal D$, using some per-sample performance metric (e.g.,\ intersection over union (IoU)) and, if applicable, applying some threshold, we obtain binarized \textbf{DuT} errors, defined as $e_i \in \{0, 1\}$. Each sample is either correctly ($e_i=0$) or incorrectly ($e_i=1$) predicted by the \textbf{DuT}.
Here, we slightly deviate from conventional notation by considering individual samples rather than the \textbf{DuT} inputs.
Although identical for image classification, in the case of object detection, multiple samples (i.e.,\ objects) may be present in a given input image, over which inference is performed.
Using a set of samples with individual errors $e_i$ allows us to define slices $\mathcal S\subseteq\mathcal D$ of the data and their corresponding (average) error rate $\restr{\bar{e}}{\mathcal S}$, defined as $\frac{1}{|\mathcal{S}|}\sum_{s\in\mathcal{S}}{e_s}$.
One of the goals of slice discovery methods is to find slices where $\restr{\bar{e}}{\mathcal S}$ is significantly worse than the global average $\restr{\bar{e}}{\mathcal D}$.
However, this constraint alone could be trivially satisfied by selecting all samples where $e_i=1$.
But, this, in general, would reveal no further information than the known data-points with bad performance.

As motivated in~\cref{sec:intro}, slice discovery methods aim to provide further information about DNN weaknesses by attempting to find semantically coherent slices.
This is typically achieved using some scheme that determines whether a specific sample $s_i$ belongs to the set $\mathcal S$. The scheme relies on additional information beyond what is provided solely by $e_i$.
This additional information may then be used to infer the cause of the weakness.
Several existing works~\citep{d2022spotlight, eyuboglu2022domino, jain2023distilling} base their schemes on distance measures in the latent spaces of the samples, so that the resulting clusters, i.e.,\ slices, require further interpretation.
In contrast, this work bases the information on metadata, more concretely a predefined set $\mathcal Z$ of semantic dimensions and corresponding attributes that describe the samples. For instance, in pedestrian detection, such dimensions can be fairness-related, such as ``gender'' or ``age'', with attributes like ``young'' or ``old'' but may also include other safety-relevant aspects such as ``occlusion'' or ``clothing-color''.
Such an algorithm offers direct interpretability of the slices, and selected dimensions can be aligned, e.g., with existing safety considerations in the form of Operational Design Domains (ODDs) for the respective systems. The set $\mathcal Z$ used in this work was inspired by \citet{herrmann2022using} in the context of automotive ODDs.

In~\cref{fig:workflow}, we present our algorithm, where, using inputs such as a dataset, a predefined ODD, and a \textbf{DuT}, we transform the task of finding systematic weaknesses in the unstructured data domain into a problem in the structured data domain. The algorithm is designed with modular components for adaptability. The first module handles the generation of structured metadata, while the second module applies the weak slice search algorithm to the generated structured metadata. With a structured description of the data, we can formulate slices as rules over $\mathcal Z$, e.g.,\ $\text{gender} = \text{male}\wedge \text{occlusion} = (0.9, 1.0]$. This allows for a more probabilistic notation $p(e|\mathcal{S})$ of the expected error given the slice. Slice discovery is then the task of finding (coherent) conditions $\mathcal S$ such that the conditional expectation is maximized.

\textbf{Metadata Generation}:  While there is great interest from safety experts and certification bodies in ODDs for safety argumentation, metadata that align with the ODD are scarcely available for most, particularly image, domains.
Human annotation of such metadata is often out of scope for large datasets due to cost and time constraints.
However, an automated metadata generation module that captures different semantic dimensions of $\mathcal{Z}$ is feasible with existing technologies. 
For example, a multi-modal foundational model like CLIP~\citep{radford2021learning} with its joint image and text embedding space could be a potential candidate for such automated annotation out of the box or after fine-tuning. 
For a given attribute $a$ we can use CLIP as a zero-shot classification function $\mathcal G$, which maps a given sample $s_i$ onto the attributes, which represent the potential classes. As such, it therefore provides the coherence of the slices discussed above.

Taking the ontology for pedestrians from the automotive domain as a baseline, a qualitative evaluation of CLIP's capability was performed by~\citet{gannamaneni2023investigating}.
While CLIP achieved SOTA level zero-shot performance on different dimensions such as gender, skin-color, and age for portrait shots of human faces in the celebA~\citep{liu2015faceattributes} dataset, they observed a drop in performance on real-world datasets containing pedestrians like in the Cityscapes~\citep{cordts2016cityscapes} dataset.
The drop in performance can be attributed to more challenging conditions, such as complex poses, low illumination, and high occlusion. 
These observations, along with our experiments, show that the classification function $\mathcal G$ is subject to varying degrees and types of uncertainty, depending on the dimensions of $\mathcal Z$:
(i) the presence of data-based (aleatoric) uncertainties, i.e.,\ where the image resolution is low or the object in question is heavily occluded or distant, leading to errors in the generated metadata. (ii) the presence of model-based (epistemic) uncertainties, i.e.,\ where the function $\mathcal G$ exhibits suboptimal performance. While (i) can occur in the case of both human and CLIP-based annotation, (ii) occurs more prominently in non-human, automated labeling.\footnote{High-quality human labeling typically requires multiple measures to reduce inter-observer variability or epistemic uncertainty in general (e.g.\ via labeling guides). However, in this work, we consider human labeling as high-quality compared to DNN-based labeling.} Therefore, any method that aims to consider metadata generated using such techniques should take into account the incurred noise in downstream tasks.

\textbf{Bayesian Framework to Account for
the Impact of Noise}: To address the uncertain nature of classification, we extend the previous slice notation of the error to the joint probability $p(e,\mathcal C, \mathcal S)$, where $\mathcal C$ represents the outcome of automated labeling for some attribute of a dimension, while $\mathcal S$ denotes the corresponding ground truth. 
For simplicity, we drop the indices and make the additional assumption that $\mathcal S,\mathcal C$ can be seen as binary, i.e.\ they may either be true ($\mathcal S$, $\mathcal C$) or not true ($\neg\mathcal S$, $\neg\mathcal C$), respectively (for details on the non-binary case, see~\cref{appendix:level1_precisions}).
Using Bayes' Theorem and marginalizing over $\mathcal C$ or $\mathcal S$, we can express
\begin{align}
    p(e|\mathcal{S}) = & p(e|\mathcal C, \mathcal S)r_\mathcal C + p(e|\neg\mathcal C, \mathcal S) \,(1-r_\mathcal C)\,,
    \label{eq:pEgivenS}
    \\
    p(e|\mathcal C)=&p(e|\mathcal C,\mathcal S)p_\mathcal C+p(e|\mathcal C,\neg \mathcal S)\,(1-p_\mathcal C)\,.
    \label{eq:pEgivenC}
\end{align}
Here, $p(e|\mathcal{S})$ represents the true slice error, while $p(e|\mathcal{C})$ denotes the observed slice error. Furthermore, $p_\mathcal C = p(\mathcal S|\mathcal C)$ and $r_\mathcal C= p(\mathcal C|\mathcal S)$ are shorthand for precision and recall of the labeling function $\mathcal G$ measured towards the ground truth, and are used in our algorithm,~\cref{fig:workflow}, for the quality check.
A detailed derivation of the equations is provided in~\cref{appendix:derivation_1}.
Making these relations explicit allows us to investigate the hypothesis typically underlying Slice Discovery Methods in more detail. Specifically, based solely on the observed slice performance/weakness $p(e|\mathcal C)$, one may conclude that a related data property $\mathcal S$ represents a weakness of the \textbf{DuT}, i.e.,\ we assume that $p(e|\mathcal S)$ also has a comparable performance/weakness. While in our algorithm the relation between $\mathcal S$ and $\mathcal C$ is explicit as the latter is given by a classifier for the former, in other approaches~\citep{d2022spotlight, eyuboglu2022domino, jain2023distilling} the relation is implicit, as observed sets $\mathcal C$ are interpreted to indicate a meaning of $\mathcal S$ (typically referred to as a slice label).
Another assumption typically made is the independence between the labeling function $\mathcal G$ and \textbf{DuT}. This independence would imply that the errors of the DuT do not depend on the noise (errors) of $\mathcal G$. Specifically, for a semantic attribute, the error rates $p(e|\mathcal C, \mathcal S)$ when $\mathcal G$ is correct and the error rate $p(e|\neg\mathcal C,\mathcal S)$ when it is not should be (approximately) equal. However, our experiments indicate that this is not always the case;
therefore, we denote the difference by
\begin{equation}
    \delta p(e|\mathcal S) = p(e|\neg\mathcal C, \mathcal S)-p(e|\mathcal C, \mathcal S)\,.
\end{equation}

Please note that $\delta p$ describes intra-set variances of the error rate in the set $\mathcal S$ and is not a conditional probability on its own.
Taking into account this potential dependence, we can derive the true error from the observed error exactly given the performance of the annotation process using
\begin{equation}\label{eq:correction_equation}
    p(e|\mathcal S) = \underbrace{\frac{p(e|\mathcal C)\,p_{\neg\mathcal C}+p(e|\neg\mathcal C)\,(p_\mathcal C-1)}{p_\mathcal C + p_{\neg\mathcal C}-1}}_\text{independence assumption}
    +\underbrace{\delta p(e|\mathcal S) \overbrace{\left(\frac{p_\mathcal C p_{\neg\mathcal C}}{p_\mathcal C + p_{\neg\mathcal C}-1}-r_\mathcal C\right)}^{\kappa_\mathcal S} 
    +\delta p(e|\neg\mathcal S)\overbrace{\frac{(p_\mathcal C -1)p_{\neg\mathcal C}}{p_\mathcal C + p_{\neg\mathcal C}-1}}^{\kappa_{\neg\mathcal S}}}_\text{correction terms}\,.
\end{equation}
As long as the independence assumption is (approximately) valid, implying $\delta p(e|\mathcal S)\approx \delta p(e|\neg\mathcal S)\approx 0$, the slice error given the semantic attribute $\mathcal S$ is obtained by separating the two types of observed error probabilities $p(e|\mathcal C)$, which is possible as long as the denominator is non-zero.\footnote{For the sake of numeric stability, also denominators which are only approximately zero should be discarded.}
An analysis of properties of this equation w.r.t.\ the denominator allows us to automatically create quality indicators on the validity or invalidity of the obtained corrected slices for attribute $\mathcal S$. The full derivation and further details on quality indicators can be found in~\cref{appendix:derivation_2}.

\textbf{Weak Slice discovery on Structured ODD Data with SliceLine}: We have now established methods to generate metadata and correct for noise during the metadata generation.
With this background, in~\cref{algo:combined_sliceline}, we propose three-stages for Systematic Weakness Detection (SWD-1,2,3). In SWD-1, using the generated structured metadata and observed errors $p(e|\mathcal C)$, we employ algorithms such as SliceLine~\citep{sagadeeva2021sliceline} to provide a ranked list of top-$k$ worst performing slices based on a scoring function that takes into account the errors and sizes of the slices (see~\cref{eq:scoring_function_orig} in~\cref{appendix:sliceline_workflow} for details on how SliceLine works).
As we have motivated, observed errors may not always provide a sufficient signal to identify the underlying error (see the top row in~\cref{fig:synthetic_data}). Therefore, in SWD-2, using~\cref{eq:correction_equation} to compensate for noise in the metadata, we provide corrected errors instead of observed errors to SliceLine to provide a second ranked list of top-$k$ worst-performing slices $\mathcal{S}$. However, as it requires extensive human effort to identify certain parameters, i.e., $\delta p(e|\mathcal S)$,\ $\delta p(e|\neg\mathcal S)$ in~\cref{eq:correction_equation}, in particular for combinations of semantics, we make a cheaper approximation only considering the independence assumption part of the equation. This is implemented in \texttt{computeCorrectedError()} in \cref{algo:combined_sliceline}. To operationalize this part of the equation, we estimate precision values based on human evaluation of metadata quality on only $n=60$ samples per attribute (see~\cref{appendix:level1_precisions}). The subsequent corrected errors from this independence assumption are used in the SliceLine scoring function.
Based on the slice quality indicators discussed above, we are also able to discard invalid slices due to denominator values close to zero.
In addition to SWD-1 and SWD-2, we also consider a merge of the resulting slices from SWD-1 and SWD-2, as this might provide a complementary effect. We refer to this merged list as the output of SWD-3. The merge step includes sorting based on the score of the slice from the scoring function, removal of duplicate slices, and filtering of invalid slices. 
The SliceLine hyperparameters include the level (maximal search depth), i.e. the maximal number of semantic dimensions considered simultaneously, as well as a cut-off for the necessary slice error $\restr{\bar{e}}{\mathcal S}$ to consider $\mathcal S$ a valid slice.

\IncMargin{1em}
\begin{algorithm}[H]
\caption{Systematic Weakness Detector (SWD)}\label{algo:combined_sliceline}
\KwIn{Metadata $\{\mathcal{C}_{\mathcal Z_1}, \mathcal{C}_{\mathcal Z_2}, \dots\}$, errors $e$, Precision vectors $\{p_{\mathcal{C}}\}$, SliceLine hyper-parameters}
\KwOut{Top-K slices $TS$}

\textbf{SWD-1: SliceLine with observed errors $p(e|\mathcal{C}_i)$}\;
\Indp
    $[TS_1] \gets \text{SliceLine}(\{\mathcal{C}_{\mathcal Z_1}, \mathcal{C}_{\mathcal Z_2}, \dots \}, e, \text{hyperparameters})$\; 
\Indm  
\textbf{SWD-2: SliceLine with corrected errors (approximations to $p(e|\mathcal{S}_i)$)}\;
\Indp
 $[TS_2, \text{Quality Indicators}] \gets \text{SliceLine}(\{\mathcal{C}_{\mathcal Z_1}, \mathcal{C}_{\mathcal Z_2}, \dots\}, \textbf{computeCorrectedError}(e, p_{\mathcal{C}}) , \text{hyperparameters})$\;
\Indm  
\textbf{SWD-3: Combined Slices}\;
\Indp
    $[TS] \gets \text{Merge}(TS_1 \cup TS_2)$\; 
\Indm

\Return TS\;
\end{algorithm}
\DecMargin{1em}

\section{Proof of Concept with Synthetic Data}
\label{sec:synthetic_data}

To demonstrate the efficacy of our algorithm~\cref{fig:workflow}, we first present evaluations on a synthetic dataset. This is done to evaluate the impact of noise on the labeling process and to determine the degree to which our algorithm can compensate for it.  
The synthetic data is a tabular dataset containing columns for nine ``real'' semantic dimensions for $200\,000$ samples each containing binary attributes. For each of the ``real'' dimensions (GT), a ``predicted'' metadata column is included as a proxy for the metadata that would be generated by CLIP in our algorithm (see~\cref{fig:workflow}). In addition, one final column contains the binarized \textbf{DuT} errors ($e$). The first four dimensions are generated to be imbalanced with only $\sim5\%$ of the samples belonging to the attribute ``1''. The other five dimensions are generated such that both attributes have equal distribution. The error column is designed such that weak slices are induced for the specified ground-truth attributes.

We consider three regimes of noise, i.e., different quality of labeling of the simulated annotation process:
(i) a regime of ``good'' quality CLIP labeling, represented with $p_\mathcal C$ above $80\%$, (ii) a regime of ``medium'' quality CLIP labeling, represented with $p_\mathcal C$ between $40\%$ and $70\%$, and (iii) a regime of ``bad'' quality CLIP labeling, represented with $p_\mathcal C$ between $10\%$ and $40\%$.
For all three regimes, we considered 100 runs of the experiments to account for statistical influence. Further details about the dataset generation can be found in~\cref{appendix:synthetic_data}.
In~\cref{fig:synthetic_data}, in the top row, the error distributions show how labeling quality impacts the spread of error between attributes for each semantic dimension, i.e.,\ the upper and lower ends of the bars are given by the error rates for $\restr{\bar{e}}{\mathcal S}$, $\restr{\bar{e}}{\neg\mathcal S}$ and similarly using $\mathcal C$ or the corrected errors. In the good labeling quality regime, as expected, observed errors and corrected errors both display the same spread as the GT error. But when labeling quality is medium or bad (where the impact of ~\cref{eq:correction_equation} is stronger), the spread of the observed error is significantly lower than that of the corrected error. In contrast, the corrected error either has close estimates to the true error or overestimates the true error (GT). From a safety perspective, we argue that overestimating the error within a DNN is better than underestimating it. In the bottom row, we evaluate the results of SWD-1,2,3. This is shown by comparing how well SWD-1,2,3 recover the top-$k$ weak slices in comparison to top-$k$ slices from Oracle, i.e., a situation where we have access to perfect ``GT'' labeling quality annotation.
Precision and recall are calculated for the three data quality regimes w.r.t.\ the Oracle case by considering the overlap of identified weak slices at an increasing number of top-slices $k$. Note that precision and recall in this figure refer to quality metrics on weak slice discovery and not precision and recall of the CLIP labeling.
While, at level 2, the maximum number of slices $k$ is 162 for 9 binarized dimensions\footnote{$9 \times 2 + \binom{9}{2}\times{2^2}$}, we consider only slices fulfilling the cut-off requirement as a weak slice. Of these 162 slices, only $\sim 30$ are identified as weak slices. Although under good labeling quality, the slices identified by SWD-1,2,3 basically have 100\% overlap with the slices from the Oracle, under medium and strong label noise, SWD-3 shows significantly more recall than SWD-1 and marginally over SWD-2. However, this comes with a small loss in precision. 
In cases of strong noise, SWD-1 only recovers a few slices where the error signal is dominant, which explains the high precision at the cost of low recall. SWD-3, on the other hand, has a reduced precision, but recovers most of the weak slices identified by Oracle. For the rest of this work, we focus primarily on the slices identified by SWD-3.

\begin{figure}[htbp]
    \centering
    \begin{tikzpicture}
        Define spacing
        \def\figwidth{5cm}
        \def\figheight{3cm}
        \def\hspace{1cm}
        \def\vspace{-3.5cm}

        \node[inner sep=0pt, outer sep=0pt] (fig1) at (0, 0.0)
            {\includegraphics[width=0.30\textwidth, height=2.5cm]{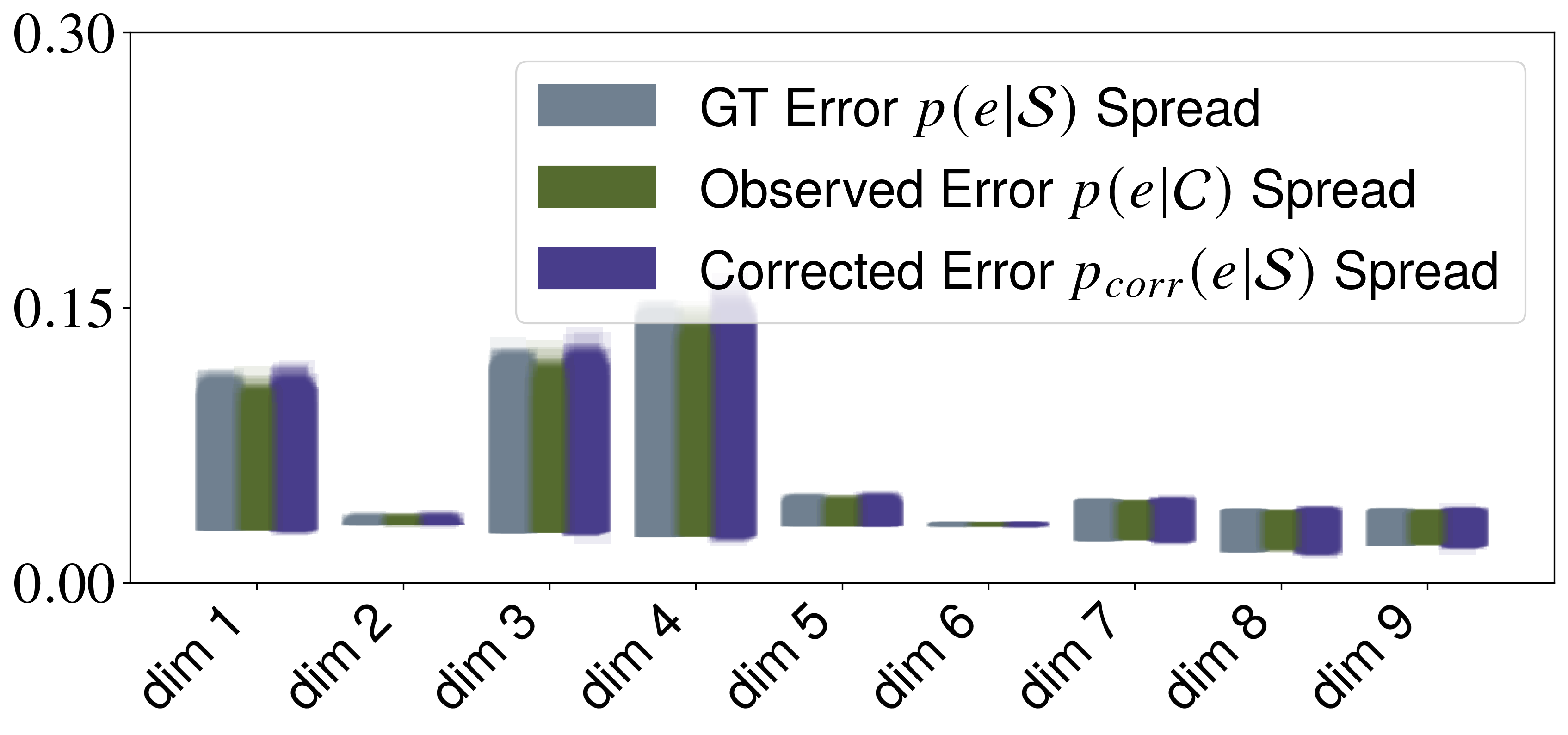}};
        \node[above=0.12cm of fig1] {Good Quality};

        \node[inner sep=0pt, outer sep=0pt] (fig2) at (\figwidth , -0.02)
            {\includegraphics[width=0.30\textwidth, height=2.42cm]{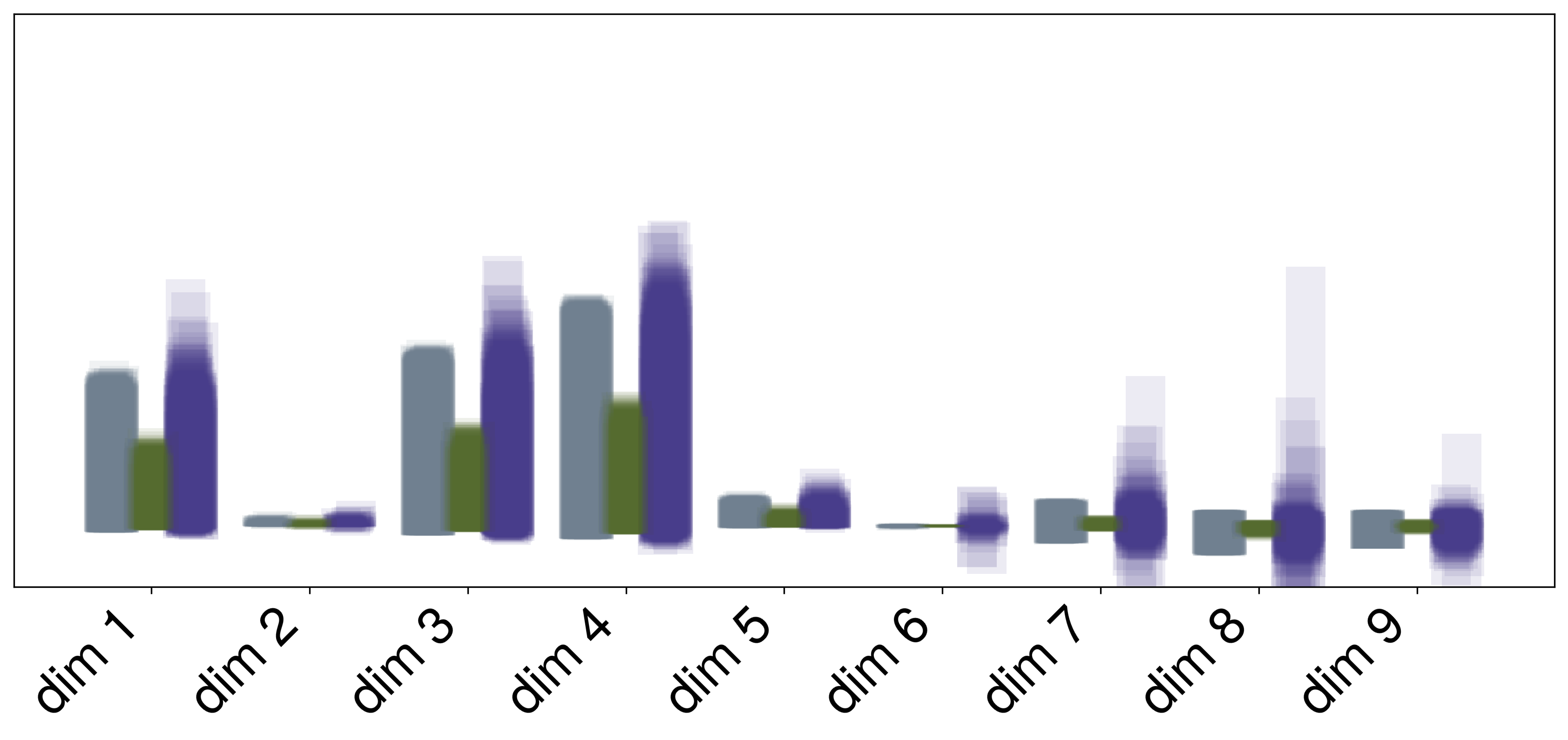}};
        \node[below=0.2cm of fig2] {Dimensions};
        \node[above=0.2cm of fig2] {Medium Quality};

        \node[inner sep=0pt, outer sep=0pt] (fig3) at (2 * \figwidth, -0.02)
            {\includegraphics[width=0.30\textwidth, height=2.42cm]{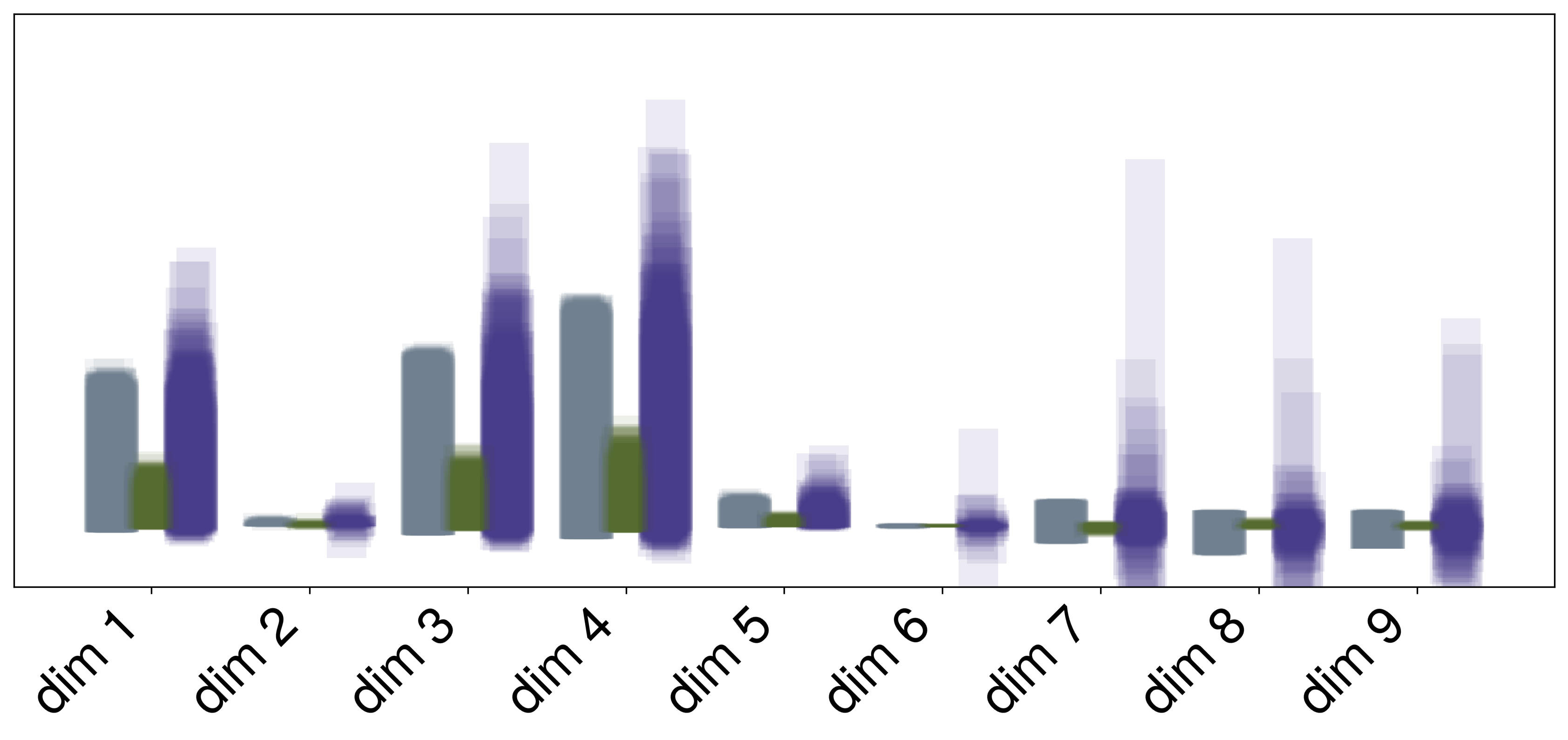}};
        \node[above=0.2cm of fig3] {Bad Quality};

        \node[inner sep=0pt] (fig4) at (0, \vspace)
            {\includegraphics[width=0.31\textwidth]{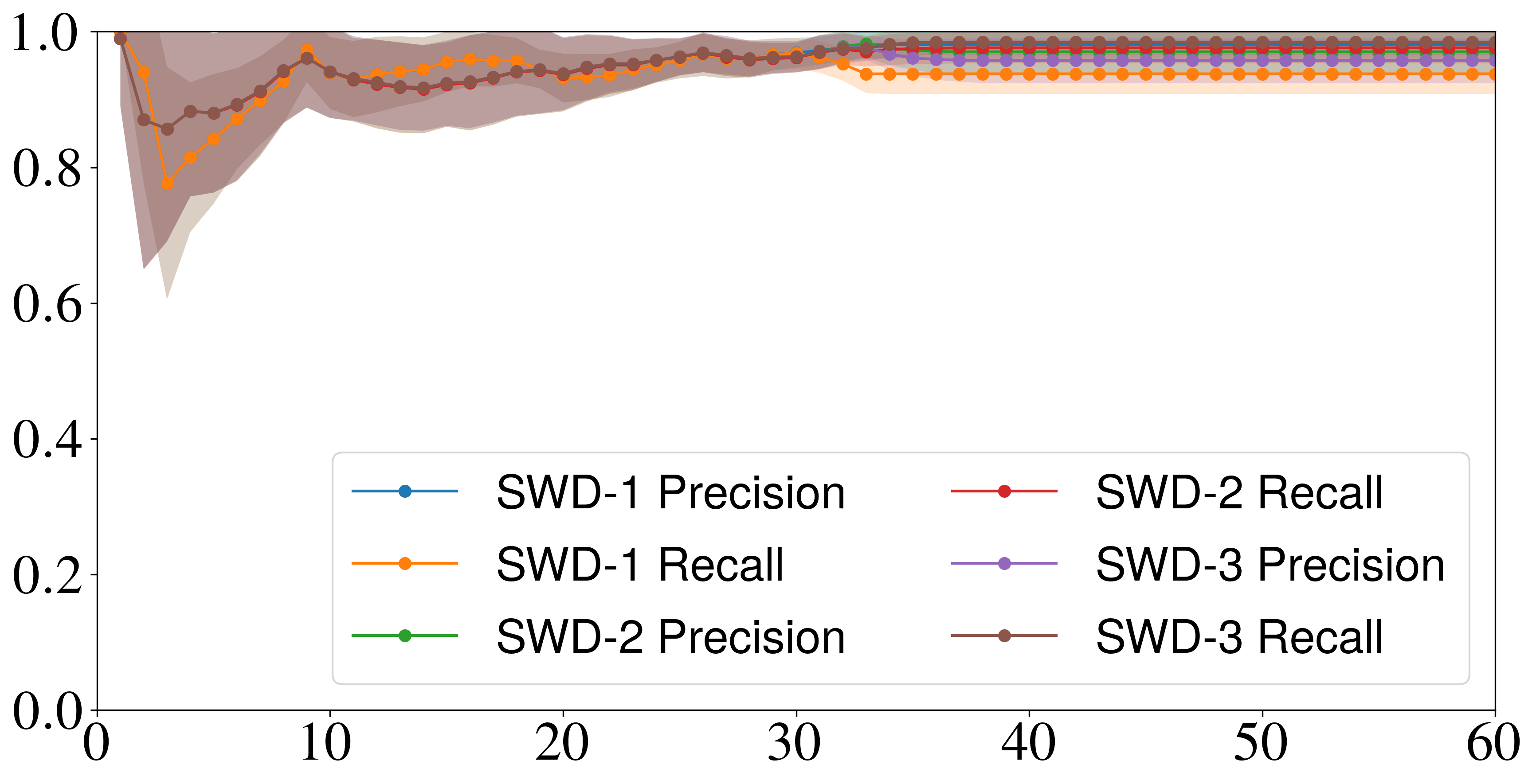}};

        \node[inner sep=0pt] (fig5) at (\figwidth, \vspace)
            {\includegraphics[width=0.31\textwidth]{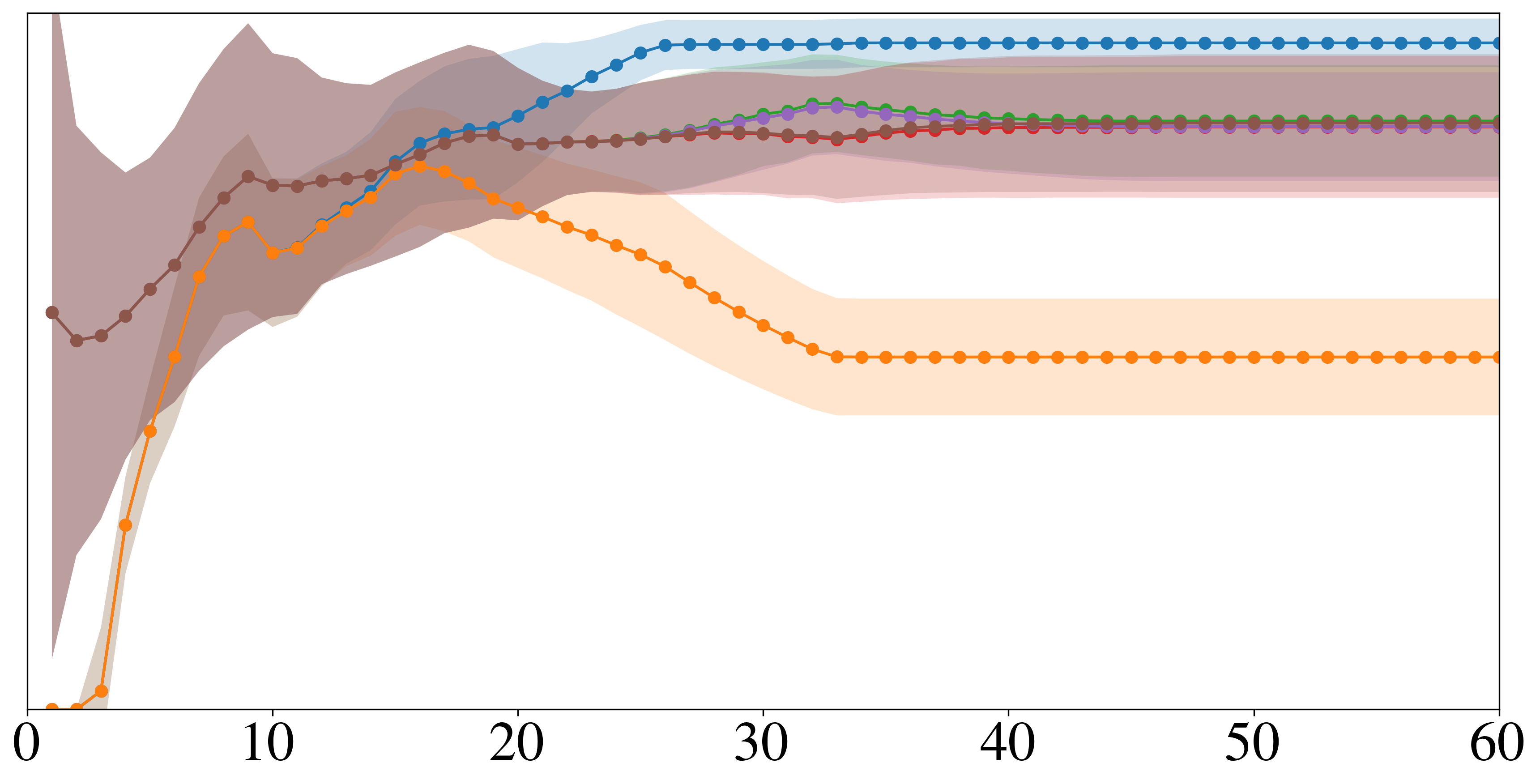}};
        \node[below=0.2cm of fig5] {Number of slices under consideration (k)};

        \node[inner sep=0pt] (fig6) at (2 * \figwidth, \vspace)
            {\includegraphics[width=0.31\textwidth]{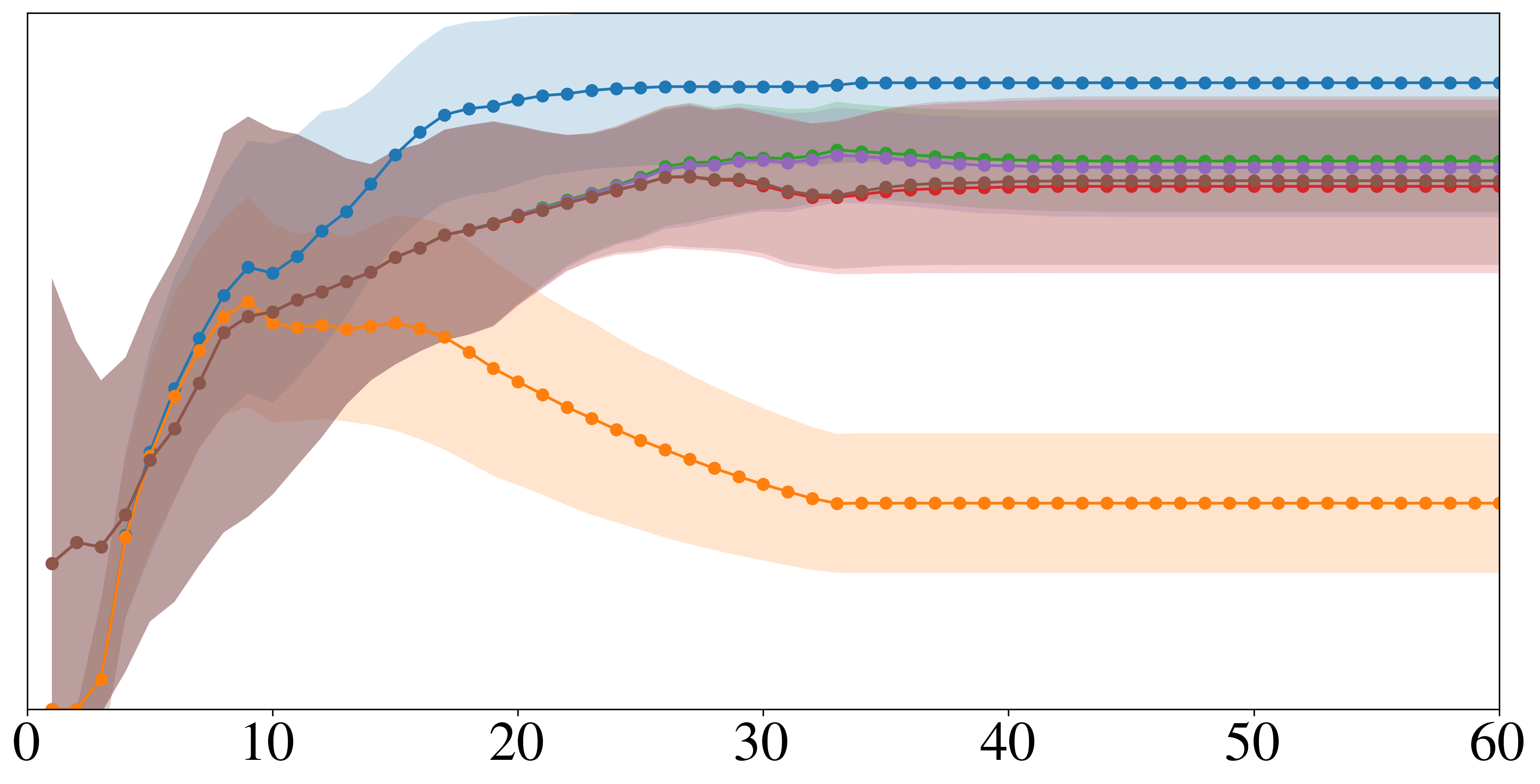}};

        \node[rotate=90] at (-2.8cm, 0.1cm) {Errors};
        \node[rotate=90] at (-3cm, -3.4cm) {\parbox{2.3cm}{\centering Precision\\Recall}};

    \end{tikzpicture}
    \caption{Based on labeling quality, we divide the generated datasets into (i) good quality (left), (ii) medium quality (middle), and (iii) bad quality (right). In three cases, we look at the spread of error in GT ($p(e|\mathcal{S})$), Observed ($p(e|\mathcal{C})$), and Corrected ($p(e|\mathcal{S})$). In the second row, corresponding performance in terms of precision and recall of SWD-1,2,3 are shown. Precision and Recall in this figure are metrics to evaluate weak slice recovery and are not related to labeling quality. The legend for both rows are presented on the figures on left.}
    \label{fig:synthetic_data}
\end{figure}
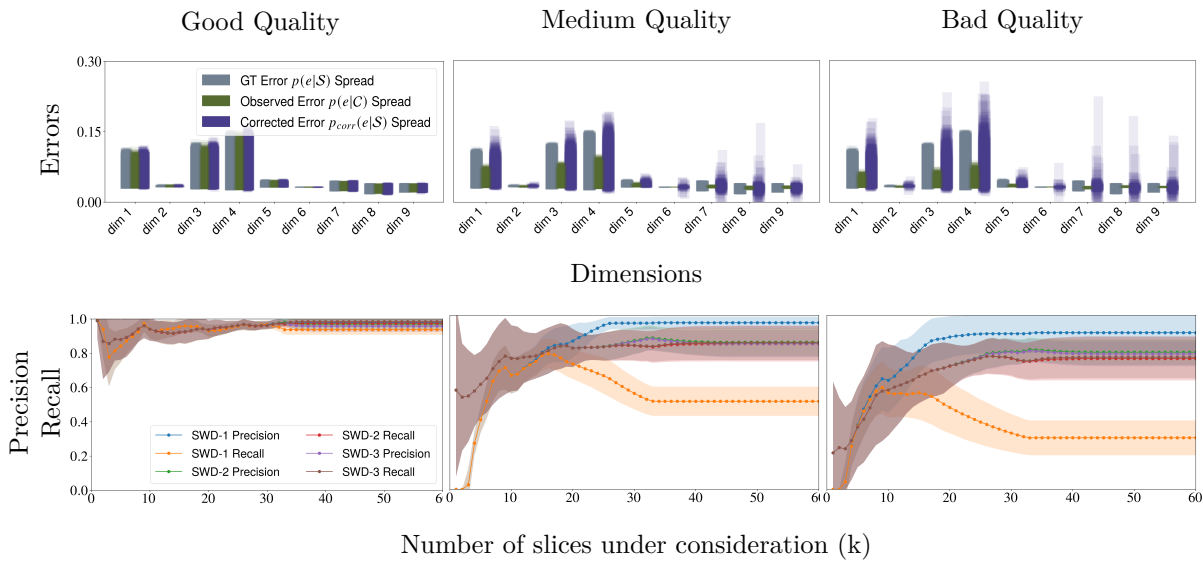

\section{Evaluations of real-world DNNs}
\label{sec:results}

In this section, we first present our experimental setup. We then show the evaluation of our systematic weaknesses detection method on a publicly available pre-trained model for the CelebA dataset. 
Here, the dataset's rich metadata annotation allows us to investigate the influence of noisy metadata annotation. In addition, we
compare against SOTA SDM methods to evaluate our claim that adherence to the ODD descriptions is useful to end users (e.g., safety experts, ML developers). Subsequently, we present the insights gained by using our approach on DNNs trained on autonomous driving datasets.

\subsection{Experimental Setup}
\textbf{Datasets and Models:} Four pre-trained models, ViT-B-16~\citep{dosovitskiy2020image}\footnote{https://github.com/huggingface/pytorch-image-models}, Faster R-CNN~\citep{ren2015faster}\footnote{https://github.com/SysCV/bdd100k-models/tree/main/det}, SETR PUP~\citep{zheng2021rethinking}\footnote{https://github.com/open-mmlab/mmsegmentation}, PanopticFCN~\citep{li2021fully} are evaluated using four public datasets (CelebA~\citep{liu2015faceattributes}, BDD100k~\citep{yu2020bdd100k}, Cityscapes~\citep{cordts2016cityscapes}, and RailSem19~\citep{Zendel_2019_CVPR_Workshops}), respectively. 
We restrict the number of combinations (level) to 2 in this work. However, as presented in~\cref{appendix:level1_precisions}, our approach allows correction of errors even at higher levels of combinations. We used the cutoff for the slice error as $1.5 \, \restr{\bar{e}}{\mathcal D}$ for all experiments except the PanopticFCN model evaluation. In the PanopticFCN evaluation, we utilize the cut-off point for the slice error as $1.0 \, \restr{\bar{e}}{\mathcal D}$ as the global average error is already quite high. For a detailed experimental setup, see~\cref{appendix:experiment_setup}. To foster reproducibility, code and the prompts used for metadata generation with CLIP will be provided.

\subsection{Evaluation of our Systematic Weaknesses Detection Method}
\label{sec:results:celebA}
\textbf{Evaluating a ViT Model on CelebA:}
As our first experiment, we evaluated the weaknesses of the ViT-B-16~\citep{dosovitskiy2020image} model (\textbf{DuT}) trained on ImageNet21k~\citep{ridnik2021imagenet}. We use the model for the targeted task of identifying the class ``person'' in the CelebA dataset~\citep{liu2015faceattributes} as a real-world proof of concept for our approach. 
Due to the extensive range of label categories in ImageNet~\citep{deng2009imagenet} and the significant noise in the labeling style, models trained on the full ImageNet dataset or its standard subset ImageNet1k~\citep{russakovsky2015imagenet} can suffer from systematic weaknesses.
For example, although the primary foreground object in an image might be a human, in some instances the image can be labeled as belonging to the class ``person'' while in other similar instances the label might be about more granular classes like ``bride'' or ``guitarist''.
To fix this issue,~\citep{ridnik2021imagenet} proposed 11 hierarchies based on WordNet~\citep{miller1995wordnet} semantic trees such that classes at higher hierarchy levels are superclasses that subsume classes at lower hierarchy levels.
However, despite these efforts, considerable label noise in terms of class overlap still persists. For example, humans holding specific objects might occur at the same hierarchy level as the class ``artifact'' or ``person''. Similar problems exist, for example, for hairstyles (see ``pompadour'' existing at the same level as ``person''). 
For a further analysis, also see the work of \citep{northcutt2021confident}.

\begin{figure}
\begin{minipage}{0.45\textwidth}
  \includegraphics[width=\linewidth]{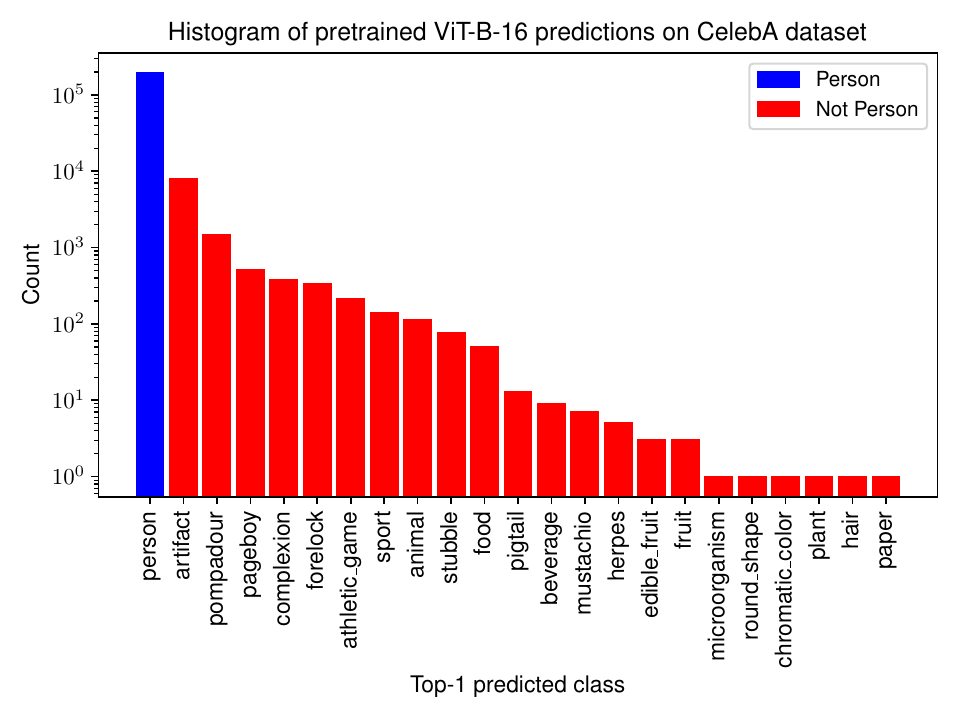} 
\end{minipage}%
\hfill 
\begin{minipage}{0.45\textwidth}
  \includegraphics[width=\linewidth]{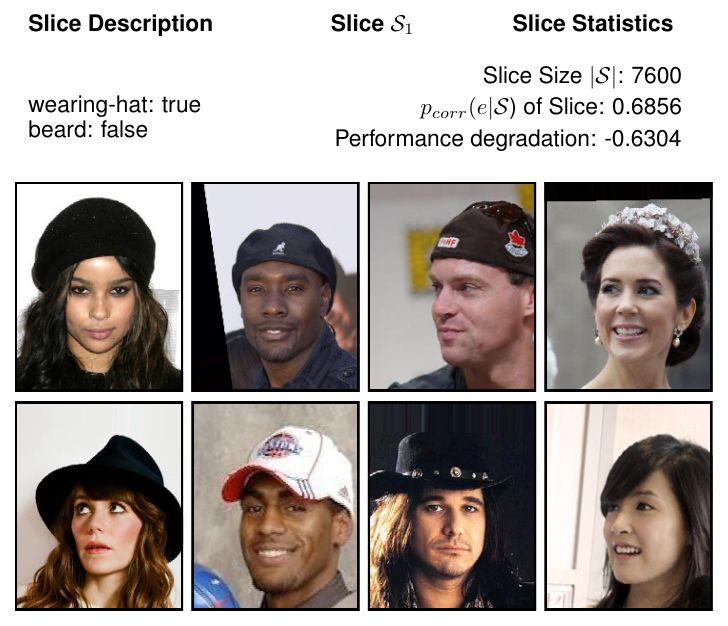} 
  
\end{minipage}
\caption{Left: The hierarchy level-0~\citep{ridnik2021imagenet} predictions of the pre-trained ViT-B-16 model on the full CelebA dataset converted into a binary classification problem. While a majority of the predictions are correct, there is a non-trivial subset of images with systematic errors due to label overlap issues. Right: Top-1 weak slice, identified by SWD-3, of a ViT-B-16 classification model trained on ImageNet21k and evaluated on the full celebA dataset. The statistics provide a quantitative evaluation of the entire slice. For qualitative evaluation, we provide some sample images from the slice.}
\label{fig:celebA}
\end{figure}

Earlier works~\citep{beyer2020we, shankar2020evaluating} have proposed using multi-label evaluation metrics as a way to deal with label noise. However, we consider the simplified task of identifying a dedicated class, ``Person'', in a dataset with only human faces (celebA) by focusing on the top-1 class predictions for level-0 of the label hierarchy proposed in ImageNet21k.
We obtain an accuracy of $94.44\%$ on the $202\,599$ images in the CelebA dataset. The softmax of the top-1 prediction, see~\cref{fig:celebA}, shows, besides the ``person'' class, the presence of several other classes, most prominently ``artifact'' and ``pompadour''.
As this model is commonly used as a pre-trained backbone for various applications, uncovering potential shortcomings might also be beneficial for potential downstream use cases of various types.
Furthermore, the CelebA dataset serves as an ideal testing ground for approaches identifying systematic weaknesses due to the availability of the ground-truth metadata attributes.
As an ODD for this test case, we propose a simplified subset of these available metadata attributes in analogy to the work of~\citet{gannamaneni2023investigating}, for details see~\cref{appendix:odds_used}.
As proposed in our algorithm, we generate metadata using CLIP for the given ODD dimensions.
Subsequently, the generated metadata is combined with the errors of the \textbf{DuT}. 

\textbf{Weak Slice Discovery}
Since the CelebA dataset contains annotated metadata for 40 attributes, we have access to noiseless metadata which, when used with SliceLine, can be considered as the ``Oracle'' approach. In~\cref{tab:first_celebA_eval}, we present the quantitative comparison of the top-7 slices identified by SWD-3 against corresponding slices in SWD-1 and Oracle. Basically, we list the top-7 slices of SWD-3 and evaluate where these slices would be ranked by SWD-1 and Oracle and what the corresponding statistics would be to highlight the importance of error correction. 
From the slice descriptions, all the identified weak slices contain some variation of the semantic concept ``wearing hat''. 
The discovery of these slices can be seen within the context of the frequent misclassification of images as class ``artifact'' by the \textbf{DuT} (see~\cref{fig:celebA}). 
In these cases, the model likely focuses on the hats as the foreground object and predicts the class ``artifact''.
To evaluate the quality of the identified slices, we utilize the errors of the slices, i.e.,\ $p_{corr}(e|\mathcal{S})$, $p(e|\mathcal{C})$, and $p(e|\mathcal{S})$ defined in~\cref{sec:method}.
We observe, based on the rank column, that the top-6 Oracle slices are captured in top-7 SWD-3 slices. Notably, while the observed error $p(e|\mathcal{C})$ of SWD-1 underestimates the true error $p(e|\mathcal{S})$ of Oracle, SWD-3 effectively corrects this in $p_{corr}(e|\mathcal{S})$. For instance, in the third row, which corresponds to the top-ranked weak slice identified by the Oracle, the difference between the Oracle slice error and the SWD-1 slice error is 0.3, while between SWD-3 and Oracle it is only 0.07. A thorough evaluation of our approach on the top-60 slices shown in~\cref{fig:celebA_furthereval} in~\cref{appendix:celeba_further_eval} reveals that SWD-3 obtains $100\%$ recall of weak slices at the cost of a reduction in precision. Note that precision and recall in this figure refer to quality metrics on weak slice discovery and not precision and recall of the CLIP labeling. From a safety perspective, given the noisy labeling, high recall (detection of all weak slices) at the cost of some reduction in precision can be considered acceptable. Interestingly, the top-1 slice of SWD-1 (not shown in table) refers to slice description ``wearing hat: true'' and ``pale-skin: true''.~\citet{gannamaneni2023investigating} discussed the limitations of CLIP in separating the latter dimension and corresponding low performance. This high level of noise in the generated metadata leads to SWD-1 identifying ``pale-skin'' as a top-1 slice while SWD-3 effectively corrects for this by discarding the wrongly detected slice as ``invalid'' using the quality indicators (see~\cref{algo:combined_sliceline}) and hence does not identify this dimension in top-7. For a qualitative evaluation of SWD-3, the top-1 slice with sample images from the slice are available in~\cref{fig:celebA} (see~\cref{fig:appendix_celebA_clip} in~\cref{appendix:qualitative_metadata_evaluation} for a qualitative evaluation of the top-5 slices).

\begin{table}[h]
    \centering
    \renewcommand{\arraystretch}{0.9} 
    \setlength{\tabcolsep}{3pt} 
    \begin{tabular}{c@{\hskip 4pt}|c@{\hskip 4pt}|ccc@{\hskip 4pt}|ccc@{\hskip 4pt}|ccc}
        Slice & Slice Description & \multicolumn{3}{c|}{SWD-3} & \multicolumn{3}{c|}{SWD-1} & \multicolumn{3}{c}{Oracle} \\
        \multicolumn{1}{c|}{} & \multicolumn{1}{c|}{} & $\operatorname{rank}(S)$ & $|\mathcal{S}|_\text{corr}$ & $p_\text{corr}(e|\mathcal{S})$ &  $\operatorname{rank}(S)$ & $|\mathcal{S}|$ & $p(e|\mathcal{C})$ & $\operatorname{rank}(S)$ & $|\mathcal{S}|$ & $p(e|\mathcal{S})$  \\
        \hline
        $\mathcal{S}_1$ & \messagebubble{\makecell[l]{\textbf{Wearing-Hat}: True \\ \textbf{Beard}: False}} & 1 & 7600 & 0.69 & 6 & 12152 & 0.33 & 2 & 6267 & 0.51 \\
        \hline
        $\mathcal{S}_2$ & \messagebubble{\makecell[l]{\textbf{Wearing-Hat}: True \\ \textbf{Smiling}: False}} & 2 & 5132 & 0.60 & 3 & 8573 & 0.36 & 9 & 6476 & 0.45 \\
        \hline
        $\mathcal{S}_3$ & \messagebubble{\makecell[l]{\textbf{Wearing-Hat}: True \\ \textbf{Gender}: Female}} & 3 & 4435 & 0.61 & 2 & 7393 & 0.38 & 1 & 2947 & 0.69 \\
        \hline
        $\mathcal{S}_4$ & \messagebubble{\makecell[l]{\textbf{Wearing-Hat}: True \\ \textbf{Age}: Young}} & 4 & 7974 & 0.54 & 4 & 12758 & 0.34 & 3 & 6937 & 0.50 \\
        \hline
        $\mathcal{S}_5$ & \messagebubble{\makecell[l]{\textbf{Wearing-Hat}: True \\ \textbf{Eyeglasses}: False}} & 5 & 8606 & 0.54 & 5 & 12594 & 0.33 & 6 & 8417 & 0.45 \\
        \hline
        $\mathcal{S}_6$ & \messagebubble{\makecell[l]{\textbf{Wearing-Hat}: True \\ \textbf{Goatee}: False}} & 6 & 8845 & 0.53 & 7 & 11453 & 0.33 & 4 & 8284 & 0.46 \\
        \hline
        $\mathcal{S}_7$ & \messagebubble{\makecell[l]{\textbf{Wearing-Hat}: True \\ \textbf{Bald}: False}} & 7 & 9676 & 0.51 & 8 & 15501 & 0.32 & 5 & 9795 & 0.44 \\
    \end{tabular}
    \caption{Evaluation of top-7 slices of SWD-3 (see~\cref{algo:combined_sliceline}) by comparing its statistics with corresponding slice statistics of SWD-1 and Oracle. The rank column indicates the slice ranking in each approach. }
    \label{tab:first_celebA_eval}
\end{table}

\textbf{Comparison to SOTA SDM method}
In addition to the evaluation of SWD-3, we compare three SOTA methods DOMINO~\citep{eyuboglu2022domino}, Spotlight~\citep{d2022spotlight}, and SVM FD~\citep{jain2023distilling} against Oracle.
Similarly to our work, DOMINO and SVM FD use CLIP (ViT L/14) in their workflows. However, they encode the images in the CLIP embedding space and then search for weak slices without explicitly enforcing any semantic concepts. To describe the slices, both approaches perform an additional step, where the identified slices are explained using text from large language models. In contrast, Spotlight directly uses the embedding space of the \textbf{DuT} to cluster weak slices and provides no descriptions of the identified slices. The former methods follow a broader trend (like us) of using foundational models like CLIP in testing smaller models. However, they do not thoroughly address limitations in CLIP's capabilities and the limitations of their approaches w.r.t.\ actionability when the slice descriptions are not very meaningful. Our approach tackles both these limitations as we address noise in CLIP labeling and also correctness of descriptions. However, our approach also has a limitation as it can only identify weaknesses w.r.t.\ dimensions in $\mathcal Z$ while the other methods could identify more novel weaknesses. However, this advantage of SOTA methods, as will be shown below, can only be realized if the description or coherence of a slice is understandable and actionable to the end-user. To assess the actionability of the SOTA methods, we consider (i) slice descriptions based on the methods themselves, (ii) slice coherence based on human inspection, and (iii) slice coherence based on overlap with top-5 slices of Oracle.

Qualitative results and slice descriptions are provided for the three methods in~\cref{appendix:qualitative_metadata_evaluation}. We identified that DOMINO descriptions can be very generic and not helpful in identifying the unique attributes of a slice. This problem was also discussed in other works~\citep{jain2023distilling, gao2023adaptive}. For Spotlight, descriptions are not available as part of the method. In contrast, in SVM FD, the slice description is targeted and covers one dimension of the weak slice identified by Oracle, namely, ``wearing hat''. However, as shown earlier, the weaknesses identified from Oracle stem from the combination of semantics. Therefore, slice descriptions from the SOTA methods are not enough for actionability. 
Second, to further evaluate the coherence of the slices, we manually inspect a sample of the images from a slice to identify the semantics. Such an approach is necessary for all methods that do not provide slice descriptions. For such manual inspection to identify the coherence of the slice, we consider samples from the slice and samples from the remaining data (last column) as a form of control group. For top-1 slices of all three approaches, it is hard to determine what uniquely constitutes the top-1 slice when considering the combination of semantics. Furthermore, such an exercise is time intensive and might potentially uncover spurious patterns.

Finally, to evaluate the coherence of the slice based on overlap with Oracle slices, we present in~\cref{tab:celebA_results} the top-1 slice identified by each method, their corresponding statistics and the overlap (Jaccard Similarity Coefficient) of the top-1 slice with the top-5 Oracle slices. From the slice statistics, it can be observed that the methods recover slices with significant performance degradation and observed error $p(e|\mathcal{C})$. However, the overlap of the top-1 slices with top-5 of Oracle is quite low. This indicates that the methods might be uncovering weaknesses w.r.t.\ dimensions not present in the ODD. However, without useful descriptions, the actionability of these slices is low. Furthermore, we evaluate the overlap of the top-1 slice with a slice that is purely made up of FNs of the \textbf{DuT}. High values in this column might indicate that priority is given to identifying FNs rather than semantic coherence, as it is unlikely that all weaknesses of a DNN can be explained by one semantic concept. Therefore, grouping all FNs into one slice would be counterproductive. As DOMINO captures $64$\% of all false negatives in its top-1 slice, it is unlikely that such a slice is actionable. In contrast, Spotlight and SVM FD capture fewer FNs in the top-1 slice. Therefore, they might be capturing some form of combination of semantics. Based on these evaluations, we conclude that the SOTA methods, when integrated with improved slice description techniques, could complement our approach. However, in their current form, our approach offers greater actionability due to its inherent slice descriptions.

\begin{table}
    \centering
    \setlength{\tabcolsep}{3pt}
    \begin{tabular}{@{}c|ccc|ccccc|c@{}}
        Method  & \multicolumn{3}{c|}{Slice Statistics} & \multicolumn{5}{c|}{Slice Coherence with Attributes} \\
        \hline 
        \multicolumn{1}{c|}{} & \makecell{Perf. \\ degr.} & Size & \multicolumn{1}{c|}{} & \multicolumn{5}{c|}{\makecell{Overlap with \\ Oracle top-5 \\ slices}} & \makecell{Overlap \\ with \\ FNs}  \\

        \multicolumn{1}{c|}{} & \makecell{$\restr{\bar{e}}{\mathcal{D}}$ \\ - \\ $p(e|\mathcal{C})$} & $|\mathcal{S}_1|$ & $p(e|\mathcal{C})$ & $J(\mathcal{S}_1, \mathcal{S}^{O}_{1})$ & $J(\mathcal{S}_1,\mathcal{S}^{O}_{2})$ & $J(\mathcal{S}_1,\mathcal{S}^{O}_{3})$ & $J(\mathcal{S}_1,\mathcal{S}^{O}_{4})$ & $J(\mathcal{S}_1,\mathcal{S}^{O}_{5})$ & $\frac{|\mathcal{S}_1 \cap \mathcal{S}_{FN}|}{|\mathcal{S}_{FN}|}$ \\
        
        \hline   
        \makecell{DOMINO} & -0.5629 & 11726 & 0.6181 & 0.13  & 0.19 & 0.20 & 0.21 & 0.24 & 0.64   \\
        \hline        
        \makecell{Spotlight} & -0.8622  & 4050 & 0.9179 & 0.32 & 0.32 & 0.32 & 0.31 & 0.31 & 0.33   
        \\
        \hline        
        \makecell{SVM FD} & -0.3844 & 2642 & 0.4295 & 0.11 & 0.15 & 0.16 & 0.16 & 0.16 & 0.24    \\

    \end{tabular}
    \caption{Comparison of three metadata-free SOTA methods with top-5 slices of Oracle. $J(\mathcal S_1, \mathcal S^O_x)$ indicates the Jaccard similarity coefficient between the two slices.~\cref{appendix:qualitative_metadata_evaluation} contains samples from each slice of the SOTA methods along with slice descriptions and statistics. For overlap with the oracle slice, higher values are better. For the overlap with the FNs, low values indicate that a slice does not contain ``significant'' weaknesses or is highly specific, while high values indicate that potentially all weaknesses of the \textbf{DuT} are in one slice and it might, therefore, be too generic. This implies that in general one would expect or desire medium overlap ranges.}
    \label{tab:celebA_results}
\end{table}

\subsection{Insights on SOTA Pedestrian Detection Models}
\label{sec:results:ad_results}
Having shown the benefits of our proposed method, we evaluate a more safety-relevant task of pedestrian detection using models trained in real-world autonomous driving (AD) datasets to identify their systematic weaknesses when predicting the class ``pedestrian''. For this, we require pedestrian level performances (intersection-over-union (IoU)) and metadata.
To avoid noisy labeling in our metadata generation step, we perform some additional steps which were not required for the previous experiment. First, we cropped all pedestrians from the images and considered these crops as $\mathcal{D}$. This is done to focus the CLIP model only on pedestrians during metadata generation.\footnote{To avoid that the aspect ratio of pedestrians is changed by the CLIP pre-processing, we use padding to obtain square crops.}

Second, we calculate the pixel area of the pedestrians based on the ground truth bounding box area and use this to filter $\mathcal{D}$ by removing pedestrians that occupy small pixel areas (``smaller'' pedestrians).
Such filtering is necessary as: 
(i) Due to low resolution and high pixelation of ``smaller'' sized pedestrians, i.e., there is a high aleatoric uncertainty regarding the correct labels affecting both CLIP and human labelers in understanding the image content (e.g., to determine gender, age, etc.).
(ii) ``Smaller'' pedestrians are more likely to be farther from the ego-vehicle~\footnote{Unless if small size is due to occlusion. For BDD100k dataset, where occlusion is available as annotation, we show impact of occlusion as well} and, therefore, might be considered less safety-relevant (in terms of vehicle breaking time).
(iii) As the small size can be strongly correlated to performance (due to distance~\citep{Gannamaneni_2021_ICCV, Lyssenko_2021_CVPR} or occlusion), this signal can strongly dominate the search for systematic weaknesses by SliceLine, thus not providing any novel insights in terms of systematic weaknesses. For this reason, we remove the low-resolution ``smaller'' pedestrians to improve the quality of metadata generation and gain further novel insights about model failures w.r.t.\ more safety-relevant pedestrians.  

The metadata generation using CLIP is performed using ODDs more suitable for automotive context (see~\cref{appendix:odds_used}).
We also perform a manual evaluation of a subset of images ($n=60$) for each attribute in each dataset to evaluate the quality of the generated metadata by estimating the precision $p_\mathcal C$ and recall $r_\mathcal C$ (as discussed in~\cref{sec:method}) and show the results in~\cref{table:estimation_metadata_quality}.

\begin{table*}[htbp!]
\centering
\begin{tabular}{@{}cc|ccc|ccc@{}}
  
  \multirow{2}{*}{\makecell{Sem. \\ dim.}} & \multirow{2}{*}{Attri.} &  \multicolumn{3}{c}{Estimated Precision $p_\mathcal C$} & \multicolumn{3}{c}{Estimated Recall $r_\mathcal C$} \\ 
 & & BDD100k & Cityscapes & RailSem19 & BDD100k & Cityscapes & RailSem19\\ 
 \hline\hline
 
 \multirow{2}{*}{Age} & Adult                    & $ 0.95 \pm 0.03 $& $ 0.99 \pm 0.02 $ & $ 0.97 \pm 0.02 $           & $ 0.76 \pm 0.03$ &$ 0.70 \pm 0.02$ & $ 0.55 \pm 0.02$\\
                      & Young                      & $ 0.69 \pm 0.06 $ & $ 0.56 \pm 0.06 $ & $ 0.42 \pm 0.06 $         & $ 0.93 \pm 0.06$ &$ 0.97 \pm 0.06$ & $ 0.94 \pm 0.06$\\
  \hline
 \multirow{2}{*}{Gender} & Female                 & $ 0.84 \pm 0.05 $ & $ 0.97 \pm 0.02 $ & $ 0.85 \pm 0.04 $         & $ 0.90 \pm 0.05$ &$ 0.95 \pm 0.02$ & $ 0.87 \pm 0.04$\\
                         & Male                 & $ 0.94 \pm 0.03 $ & $ 0.97 \pm 0.02 $  & $ 0.94 \pm 0.03 $          & $ 0.88 \pm 0.03$ & $ 0.97 \pm 0.02$ & $ 0.92 \pm 0.03$\\
 \hline
 \multirow{2}{*}{\makecell{Cloth.-\\color}} & \makecell{Bright-\\color}  & $ 0.81 \pm 0.05 $ & $ 0.85 \pm 0.04 $  & $ 0.79 \pm 0.05 $         & $ 0.30 \pm 0.05$ &$ 0.23 \pm 0.04$ & $ 0.66 \pm 0.05$\\
                         & \makecell{Dark-\\color}            & $ 0.76 \pm 0.05 $ & $ 0.65 \pm 0.06 $ & $ 0.82 \pm 0.05 $         & $ 0.96 \pm 0.05$&$ 0.97 \pm 0.06$ & $ 0.89 \pm 0.05$\\
 \hline
 \multirow{2}{*}{\makecell{Skin-\\color}} & Dark             & $ 0.82 \pm 0.05 $&  $ 0.55 \pm 0.06$ & $ 0.56 \pm 0.06 $           & $ 0.92 \pm 0.05$& $ 0.71 \pm 0.06$ & $ 0.76 \pm 0.06$\\
                         & White                  & $ 0.99 \pm 0.02 $&  $ 0.95 \pm 0.03$ & $ 0.89 \pm 0.04 $           & $ 0.96 \pm 0.02$ & $ 0.91 \pm 0.03$ & $ 0.75 \pm 0.04$\\
 \hline
 \multirow{2}{*}{Blurry}                         & True &  $ 0.71 \pm 0.06 $ &  $ 0.63 \pm 0.06$ & $ 0.87 \pm 0.04$  & $ 0.42 \pm 0.06$& $ 0.87 \pm 0.06$ & $ 0.64 \pm 0.04$\\
                                                 & False  &  $ 0.48 \pm 0.06 $ &  $ 0.95 \pm 0.03$ &$ 0.84 \pm 0.05$ & $ 0.74 \pm 0.06$ & $ 0.82 \pm 0.03$ & $ 0.95 \pm 0.05$\\
                                                  \hline
 \multirow{2}{*}{\makecell{Constru.- \\ Worker}}                         & False &  - &  - & $ 0.97 \pm 0.02$  &  -& - & $ 0.98 \pm 0.02$\\
                                                 & True  &  - & - &$ 0.65 \pm 0.06$ & - & - & $ 0.55 \pm 0.06$\\
\end{tabular}
\caption{The estimated precision and recall using our proposed approach for evaluating the quality of the generated metadata. Here, we provide the mean and $\sigma/2$, for $n$ of 60, of the estimated precision and recall. Certain dimensions like occlusion are available as part of the datasets themselves. We do not perform human-evaluation for these dimensions but these are considered in the weak slice search.}
\label{table:estimation_metadata_quality}
\end{table*}

In these experiments, using SWD-3, we evaluate the weaknesses of an object detection model (Faster R-CNN), a segmentation model (SeTR PUP), and a panoptic segmentation model (Panoptic-FCN). The models are evaluated on their respective datasets, i.e., BDD100k, Cityscapes, and RailSem19. Samples of image crops of the identified top-1 weak slice for each experiment are shown in~\cref{fig:ad_results} (see~\cref{fig:appendix_bdd100k,fig:appendix_cityscapes,fig:appendix_railsem} in~\cref{appendix:qualitative_metadata_evaluation} for top-5 weak slices). In~\cref{tab:summary_ad}, we present the largest and worst performing slice of the top-5 to provide insights about the three models. In all three experiments, the performance degradation of the identified slices is significant. ``Occlusion'', skin-color and clothing-color are reoccurring slice descriptions for the first two models, which are tested on datasets that contain images with many nighttime scenes (BDD100k) or relatively high gray-toned scenes (Cityscapes). In contrast, the third model, which contains relatively brighter scenes, has a significant weakness for the dimension ``age''. The estimated precision $p_\mathcal C$ and recall $r_\mathcal C$ in~\cref{table:estimation_metadata_quality} were provided as input to~\cref{algo:combined_sliceline} to obtain these slices and to determine the quality of the identified weaknesses. Therefore, in contrast to SOTA SDMs, our approach identifies human-understandable safety-relevant systematic weaknesses in DNNs used for real-world applications.

\begin{figure}[ht]
\begin{minipage}{0.33\textwidth}
  \includegraphics[width=\linewidth]{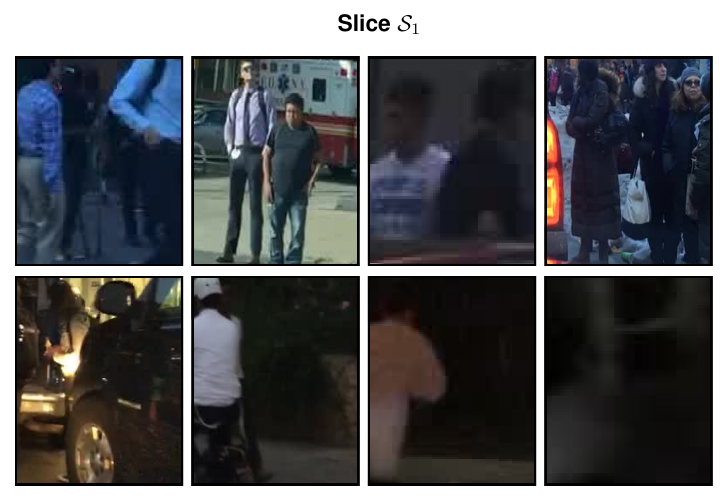} 
   \subcaption[]{Faster R-CNN}
\end{minipage}%
\begin{minipage}{0.33\textwidth}
  \includegraphics[width=\linewidth]{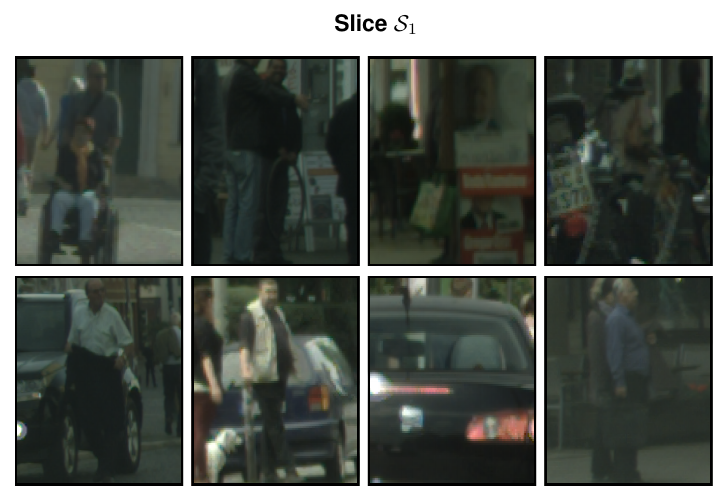} 
   \subcaption[]{SeTR}
\end{minipage}
\begin{minipage}{0.33\textwidth}
  \includegraphics[width=\linewidth]{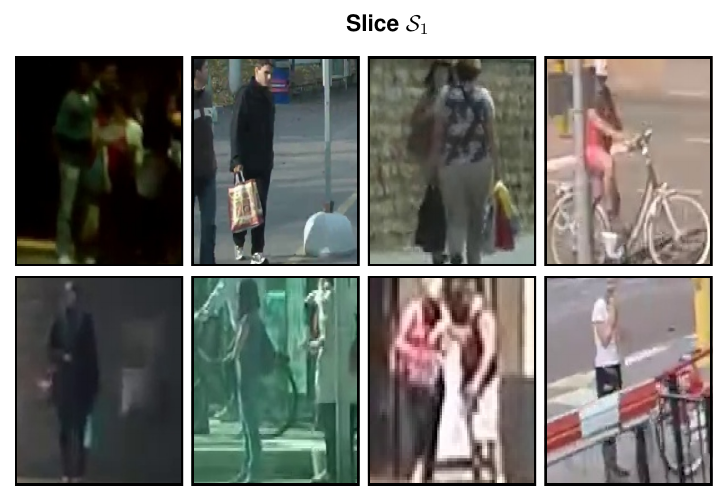} 
   \subcaption[]{Panoptic-FCN}
\end{minipage}
  \caption{Left: Samples from top-1 weak slice of a Faster R-CNN object detector trained and evaluated on BDD100k dataset. Middle: Samples from top-1 weak slice of SeTR model trained and evaluated on Cityscapes dataset. Right: Samples from top-1 weak slice of a Panoptic-FCN model trained and evaluated on RailSem19 dataset.}
\label{fig:ad_results}

\end{figure}

\begin{table}[bp!]
\setlength{\tabcolsep}{7pt}
    \centering
    \begin{tabular}{c|ccc|ccc}
        \multirow{2}{*}{\makecell{Model \&\\ Dataset}} &  \multicolumn{3}{c}{\makecell{Largest Slice \\ (in top-5)}} & \multicolumn{3}{c}{\makecell{Worst Performing Slice \\ (in top-5)}} \\
        
        & $\frac{|\mathcal{S}|}{|\mathcal{D}|}$\% & $p_\text{corr}(e|\mathcal{S})$ & \makecell{Perf. \\ Degr.} & $\frac{|\mathcal{S}|}{|\mathcal{D}|}$\% & $p_\text{corr}(e|\mathcal{S})$ & \makecell{Perf. \\ Degr.}   \\
         
        \hline 
        {\makecell{Faster R-CNN \\ BDD100k}} & 34.44\% & 0.1263 & -0.0693 &  14.22\% & 0.2206 & -0.1636 \\

        \hline 
        {\makecell{SeTR \\ Cityscapes}} & 13.34\% & 0.0594 & -0.0446 &  9.24\% & 0.1046 & -0.0897 \\ 
        
        \hline
        {\makecell{Panoptic-FCN \\ RailSem19}} & 25.49\% & 0.8663 & -0.222 &  8.13\% & 1.0 & -0.4602 \\
        
    \end{tabular}
    \caption{Quantitative analysis of three pre-trained autonomous driving models (results are only for SWD-3). From the top-5 weak slices, we show the largest slice and the weakest performing slice. Please refer to the~\cref{appendix:qualitative_metadata_evaluation} for the top-5 slices.}
    \label{tab:summary_ad}
\end{table}

\section{Conclusion}
\label{sec:conclusion}

In this work, we present an algorithm for our Systematic Weakness Detector (SWD) to analyze the systematic weaknesses of DNNs that perform classification, object detection, and semantic segmentation tasks on image data. In the first step, we overcome the problem of missing metadata by generating metadata with a foundation model. Subsequently, in the second step, we perform slice discovery on the structured metadata, which comprises of DNN-under-test's per-object performance and previously acquired per-object metadata. Using our algorithm, we transform the slice discovery of unstructured image data into an (approximate) slice discovery problem on structured data.
In addition, we study the impact of noisy labeling in a Bayesian framework and operationalize it by integrating error correction and slice validity based on quality indicators into our approach.
In the ablation experiments, we show that our SWD detects the same weak slices as would be identified in hypothetical cases where we have access to perfect metadata.
The primary advantage of our algorithm, in comparison to SOTA methods, is that the identified weak slices are aligned with human-understandable semantic concepts that can be derived from a description of the ODD.
As upcoming safety and trustworthy AI specifications require evidences for building safety argumentations w.r.t.\ such ODDs, the results from our approach can directly contribute.
In addition, the identification of human-understandable weak slices enables ML developers to take mitigation actions, such as a targeted acquisition or generation of data, addressing the weaker slices and, thus, facilitating effective re-training with a limited acquisition budget.
Furthermore, we show that our approach has clear advantages over several metadata-free SOTA methods by giving more actionable results, and we demonstrate the applicability of our approach by identifying systematic weaknesses in multiple AD datasets.
For this, we also provide a quantitative evaluation of the quality of the generated metadata. 

Our algorithm does have certain limitations. Primarily, a minimum metadata labeling quality is required for the discovered slices to be meaningful. In addition to our proposed metadata quality estimation, future works could therefore focus on improving metadata quality by human correction of a subset of generated metadata, fine-tuning~\citep{eyuboglu2022domino} of CLIP, metadata acquisition from other sources (e.g., depth sensor).
Secondly, all approaches based on ODD definitions, like ours or PromptAttack~\citep{Metzen_2023_ICCV}, would suffer from the lack of completeness of the semantic concepts in ODD. 
A potential solution could be in the direction of~\citet{gannamaneni2024assessing} by performing a root-cause analysis of found weaknesses.
Such approaches could address potential issues between correlation and causation for found small slices.
In addition, SDMs based on the evaluation of the test dataset can suffer from insufficient coverage of the application domain by the test dataset.
Both aspects become more relevant with the increasing broadness of the assumed ODD scope. For instance, if one intends to investigate false positives in object detection, the description would effectively contain most other objects (and parts thereof) that could appear in the scene. 
While we, therefore, limit our scope to the more narrowly defined false negatives, our approach still provides valuable insights into, often more critical, missed detections in terms of human-understandable and, thereby, actionable weak slices. 
We believe that such results can contribute to the development of trustworthy AI models and their safety.

\section{Acknowledgments}
This work has been funded by the European Union and the German Federal Ministry for Economic Affairs and Climate Action as part of the safe.trAIn project.
\bibliography{main}
\bibliographystyle{tmlr}

\appendix
\clearpage
\setcounter{page}{1}
\begin{center}
    \Large
    \textbf{Appendix}
   
\end{center}

\appendix

\section{Experiment Setup}
\subsection{Datasets, DNNs, Hyperparamters}
\label{appendix:experiment_setup}

\textbf{Datasets:} We use four datasets in our experiments to test different black-box DNNs-under-test. First, as a proof-of-concept for our overall approach, we use the CelebA~\citep{liu2015faceattributes} dataset with a rich collection of 40 facial attributes (metadata) for $202\,599$ images of celebrity faces. We used the aligned PNG images provided by the authors, which have a resolution of $178 \times 218$ pixels.
Next, for pedestrian detection tasks, we consider BDD100k~\citep{yu2020bdd100k}, Cityscapes~\citep{cordts2016cityscapes}, and RailSem19~\citep{Zendel_2019_CVPR_Workshops}. In these datasets, we focus only on the pedestrian class in the 2D-bounding box and semantic segmentation tasks. For BDD100k, we consider the predefined validation set of 10k samples of resolution $1280 \times 720$, while in Cityscapes and RailSem19, due to their smaller dataset sizes, we use the entire train and validation sets containing 3475 (the test set is not considered due to the lack of GT) and 8500 samples with image resolutions $2048 \times 1024$ and $1920 \times 1080$ respectively.

\textbf{Models (DNNs-under-test):} We evaluate four black-box models for ODD aligned systematic weaknesses. For the first experiment, we consider the publicly available ViT-B-16~\citep{dosovitskiy2020image} model pre-trained on ImageNet21k~\citep{ridnik2021imagenet} from the python library timm.\footnote{https://github.com/huggingface/pytorch-image-models}
Second, we use the pre-trained publicly available Faster R-CNN~\citep{ren2015faster} object detector with ConvNeXt-T~\citep{liu2022convnet} backbone. The model weights are available on the BDD100k model Zoo.\footnote{https://github.com/SysCV/bdd100k-models/tree/main/det} Third, for the Cityscapes dataset, we use a pre-trained SETR PUP~\citep{zheng2021rethinking} semantic segmentation model. The model weights are available on the mmsegmentation codebase.\footnote{https://github.com/open-mmlab/mmsegmentation}
Finally, from the railway domain, we use a PanopticFCN~\citep{li2021fully} model, which has been trained by an industrial partner on a large proprietary dataset also including RailSem19~\citep{Zendel_2019_CVPR_Workshops}. We consider this as a complete black box and have no details on the concrete training procedure. 
For the autonomous datasets, the black-box model performance per-object (i.e., pedestrian) is measured by the intersection-over-union (IoU).

\textbf{Parameters} of CLIP and SliceLine:
For metadata generation, we use a pre-trained CLIP~\citep{radford2021learning} with image encoder (ViT-L/14~\citep{dosovitskiy2020image}).
For SliceLine, we use a python implementation and choose default $\alpha$ and $\sigma$ values of $0.95$ and $n/100$ where $n$ defines the size of the structured data as proposed in \citet{sagadeeva2021sliceline}. For the synthetic data experiment, we incrementally increase k from 1 to 60. 

\subsection{Synthetic Data Generation Parameters}
\label{appendix:synthetic_data}

The purpose of the synthetic data experiment is to evaluate the algorithm with control over the quality of labeling and without the influence of correlations. Therefore, we build a tabular dataset with 9 ``real'' semantic dimensions (dim1, \dots, dim9) containing 200,000 rows. For each of these dimensions, we generate a synthetic dimension as a proxy for labeling by CLIP. All dimensions contain binary attributes. For the first five dimensions, the distribution of true attribute is imbalanced, i.e., only 5\% of overall samples ([8000, 9000, 10000, 11000, 12000]), respectively. The other dimensions are balanced between both attributes. The final column contains errors simulating the \textbf{DuT} performance. Next, we define a set of slices and induce errors for each of the slices. For our experiments, we induce the following errors: \{dim1: 0.19, dim2 \& dim3: 0.18, dim3: 0.23, dim4: 0.3, dim5: 0.07, dim6: 0.04, dim7: 0.01, dim8: 0.05, dim9: 0.02\}. As we have 100 runs for different labeling qualities, we introduce random fluctuations between -0.01 and 0.01 to these error values. The choice of errors and number of dimensions is to align the synthetic data with the celebA experiment and also to effectively induce errors. If all dimensions contribute roughly equally to the error rate, no strong signal for a specific slice, in contrast to the others, could be found. We generate 100 runs each for 3 different labeling qualities.
That is, we generate the ``observed'' metadata from the ``real'' one using a random predefined ``precision'' value.
For good quality, this precision value to detect attribute 1 of each dimension is sampled from a uniform distribution between 0.8 and 1.0. For attribute 0, we sample between 0.8 and 1.0. For medium quality and attribute 1, we sample from 0.4 and 0.6 and for attribute 0 between 0.4 and 0.7. For bad quality, for attribute 1, we sample between 0.1 and 0.4 and for attribute 0 between 0.3 and 0.6.

\subsection{Human-understandable Dimensions}
\label{appendix:odds_used}

\citet{herrmann2022using} have proposed ontologies for different dynamic objects (e.g., pedestrians) to build ODDs for AD vehicles. Although these proposed ontologies do not yet completely capture all safety-relevant features, they provide a reference to the direction safety experts intend to take to build evidences for safety augmentations of AD vehicles. To enable such a formulation of evidence, we performed our experiments on a subset of the concepts discussed in these ontologies as shown in~\cref{table:AD_ontology,table:celebA_ontology}. In the case of BDD100k, as information about occlusion is provided in the dataset, we combine our generated metadata with this additional information. For the CelebA experiment, as the input distribution is not directly related to the AD domain, we consider semantic concepts that are more suitable for this dataset as shown in~\cref{table:celebA_ontology}. Similar to~\citet{gannamaneni2023investigating}, we encode the input image using the CLIP image encoder. For celebA, we encode the entire input image, while for the AD experiments, we encode individual pedestrian crops as a single input. We consider each semantic dimension and its corresponding attributes to generate metadata for an input image. 

\begin{table}[htbp!]
\centering
\setlength{\tabcolsep}{4pt} 

\begin{tabular}{c|cc} 
 \hline
 Semantic dimension & \multicolumn{2}{c}{Attributes}   \\ 
 \hline\hline
 Gender             & Male & Female \\
 Pale-skin         & True & False \\
 Age                & Young & Adult \\
 Beard              & True & False   \\ 
 Goatee             & True & False \\
 Bald               & True & False \\
 Wearing-Hat        & True & False \\
 Wearing-Eyeglasses & True & False \\
 Smiling            & True & False \\
\hline
\end{tabular}
\caption{The ODD used for the celebA experiment. The first column represents the different semantic dimensions (in analogy to safety-relevant features). For each dimension, different attributes are considered and generated as metadata using our metadata generation process.}
\label{table:celebA_ontology}
\end{table}

\begin{table*}[htbp!]
\centering

\begin{tabular}{c|ccccc} 
 \hline
 Semantic dimension & \multicolumn{2}{c}{Attributes}   \\ 
 \hline\hline
 Gender              & Male & Female \\
 Skin color          & White & Dark \\
 Age                 & Young & Adult \\
 Clothing color      & Bright-color & Dark-color \\
 Blurry              & True & False \\
 $\text{Occlusion}^\dag$         &    True   &   False      \\ 
 $\text{Construction-worker}^{\ddag}$ & True & False \\
 Size                & 10 quantile binned values of bounding box pixel area\\ 

\hline
\end{tabular}
\caption{A sample ontology for pedestrians used in our AD dataset experiments. The first column represents the different semantic dimensions (safety relevant features). For each dimension, different attributes are considered and generated as metadata using our metadata generation process. Metadata that is generated from CLIP but from available through other sources (e.g., GT) is not considered noisy and, therefore, we do not perform precision and recall estimation by human sampling. $\dag$ Occlusion is available as GT from the BDD100k dataset and we only consider it in corresponding experiment. ${\ddag}$ RailSem19 dataset contains several images where construction-workers are present near railway tracks. Therefore, we additionally consider this dimension for CLIP labeling to identify if models have weaknesses identifying construction workers. Size of pedestrian is estimated by calculating product of bounding width and height.}
\label{table:AD_ontology}
\end{table*}

\subsection{SliceLine Workflow}
\label{appendix:sliceline_workflow}

SliceLine works on individual errors $e_i$ of data samples $i$. These, in the original work, can be defined as $e_i=1-p_i$ with the DuT predicted probability $p_i$ for the correct class. In the remainder, we make the simplifying assumption that $e_i\in\{0,1\}$ indicates whether $i$ was classified correctly, $e_i=0$, or not, $e_i=1$.
The workflow of SliceLine to identify weak slices is as follows: Initially, for depth level 1, a breadth search is performed on all attributes in the metadata such that only single features form a slice (e.g., a slice containing all data points with condition $(gender: male)$).
Checks are performed over these slices to ensure that thresholds are met (e.g., minimum slice size specified via some parameter $\sigma$). 
Next, based on the slice scores from~\cref{eq:scoring_function_orig}, the slices are ordered, and a list of top-k weak slices is populated.
The hyperparameter $\alpha$ in \cref{eq:scoring_function_orig} allows us to weight the size of the slice as well as the error signal.
At depth level 2 and above, combinations of two attributes are chosen to form a slice (e.g., slice containing all data points with condition $(gender = male) \& (occlusion = (0.9, 1.0])$).
The list of weak slices is updated after each depth level. The maximum depth level is a hyperparameter.
In addition, pruning steps are also performed at each depth level in the original implementation.  
The conditions for pruning have a monotonicity property, which ensures that all potential sub-slices of a pruned slice would also fulfill the pruning condition. 
Due to the limited sizes of the ODDs for our experiment, we do not consider the pruning step in our implementation.
Once the maximum depth level has been reached, the algorithm is terminated and the final list of top-$k$ weak slices are available.

\begin{equation}\label{eq:scoring_function_orig}
    \text{Scoring Function}(\mathcal{S}) = \alpha\,\frac{\restr{e}{\mathcal{S}}-\restr{e}{\mathcal{D}}}{\restr{e}{\mathcal{D}}} - \left(1 - \alpha\right)\,\frac{|\mathcal{D}|-|\mathcal{S}|}{{|\mathcal{S}|}}
\end{equation}

\section{Derivations}
\subsection{Derivation of \texorpdfstring{$p(e|\mathcal C)$}{p(e|C)} and  \texorpdfstring{$p(e|\mathcal S)$}{p(e|S)}}
\label{appendix:derivation_1}

To derive \cref{eq:pEgivenS} and \cref{eq:pEgivenC} from \cref{sec:method}, we first consider the joined probability $p(e,\mathcal C,\mathcal S)$, where $e$ denotes the DuT error, $\mathcal{C}$ labeling, and $\mathcal{S}$ the ground truth for some semantic attribute. Using Bayes' Theorem we can rewrite this as
\begin{equation}
    p(e,\mathcal C,\mathcal S) =
    p(e|\mathcal{C}, \mathcal S) p(\mathcal{C}, \mathcal S) =
    p(e|\mathcal{C}, \mathcal S) p(\mathcal{C}| \mathcal S) p(\mathcal S)
    \,.
\end{equation}
Looking additionally at marginal distributions
\begin{equation}
    p(e,\mathcal S) = \sum_\mathcal{C} p(e,\mathcal C, \mathcal S) 
    = p(\mathcal S) \sum_\mathcal{C} p(e|\mathcal C, \mathcal S)p(\mathcal C| \mathcal S)
    \,,
\end{equation}
where the sum goes over all possible values $\mathcal C$ can take.
We can write the conditional error probability (or rate if considered over finite data) as
\begin{equation}
    p(e|\mathcal S) = \sum_\mathcal{C} p(e|\mathcal C, \mathcal S)p(\mathcal C| \mathcal S)\,
\end{equation}
At this point, using that $\mathcal C$ takes only binary values, which, for brevity, we denote as $\mathcal C$ if the attribute was detected and as $\neg\mathcal{C}$ else,\footnote{This is a slight over-use of the notation, but it is apparent from context whether $\mathcal C$ is meant as the random variable for the labeling, or as its value in the sense of positive detection.} we can expand the sum:
\begin{equation}
    p(e|\mathcal S) = p(e|\mathcal C, \mathcal S)p(\mathcal C| \mathcal S)
    + p(e|\neg\mathcal C, \mathcal S)p(\neg\mathcal C| \mathcal S)
    \,
\end{equation}
Within this expression, we can identify the recall
\begin{equation}
    r_\mathcal{C} \equiv p(\mathcal C | \mathcal S)
\end{equation}
of the labeling method, that is the probability we will obtain correct identification of the semantic attribute given its presence.
Using further the normalisation property
\begin{equation}
    1=\sum_\mathcal{C}p(\mathcal C| \mathcal S)
    \quad\rightarrow\quad
    p(\neg\mathcal C|\mathcal S) = 1-p(\mathcal C| \mathcal S)
    \,,
\end{equation}
we arrive at the originally presented \cref{eq:pEgivenS}:
\begin{equation}
    \label{eq:prec_and_recall}
    p(e|\mathcal S) = r_\mathcal{C}\,p(e|\mathcal C, \mathcal S)
    + \left(1-r_\mathcal{C}\right)\,p(e|\neg\mathcal C, \mathcal S)
    \,
\end{equation}
Along the same lines \cref{eq:pEgivenC},
\begin{equation}
     p(e|\mathcal C) \equiv  p_\mathcal{C}\,p(e|\mathcal C, \mathcal S)
    + \left(1-p_\mathcal{C}\right)\,p(e|\mathcal C, \neg\mathcal S)
    \,,
\end{equation}
can be derived, however with the identification
\begin{equation}
    p_\mathcal{C}= p(\mathcal S|\mathcal C)\,,
\end{equation}
i.e., the precision of the labeling process.

\subsection{Derivation of  Correction Equation}
\label{appendix:derivation_2}

As discussed in~\cref{sec:method}, the annotation process may not be a perfect process. Furthermore, there is no guarantee that the failure modes within this process do not overlap the failures of \textbf{DuT}, i.e., there is a possibility that some amount of correlation could exist between the errors of annotation process and the errors of \textbf{DuT}. Therefore, we frame this using the following
\begin{equation}\label{eq:correlation_correction_term}
    \delta p(e|\mathcal S) = p(e|\neg\mathcal C, \mathcal S)-p(e|\mathcal C, \mathcal S)
\end{equation}
By considering earlier equations and their complementary forms for $\neg\mathcal S$ and reducing the equation set, we obtain

\[
A = 
\begin{pmatrix}
  p_\mathcal C & 1 - p_\mathcal C \\
  1 - p_{\neg\mathcal C} & p_{\neg\mathcal C}
\end{pmatrix},
\quad
B = 
\begin{pmatrix}
  p(e|\mathcal C) + (p_\mathcal C)\, \delta p(e|\neg\mathcal S) \\[6pt]
  p(e|\neg\mathcal C) + (p_{\neg\mathcal C})\, \delta p(e|\mathcal S)
\end{pmatrix},
\]

\[
A
\begin{pmatrix}
  p(e|\mathcal C, \mathcal S) \\[4pt]
  p(e|\neg\mathcal C, \neg\mathcal S)
\end{pmatrix}
=
B
\]

Here, the $det(A)$ is given by $p_c + p_{-c} - 1$ and inverse of $A$ is given by

\begin{equation}\label{eq:matrix_corr_equation}
A^{-1} 
= 
\frac{1}{p_c + p_{-c} - 1}
\begin{pmatrix}
  p_{-c} & -\!\bigl(1-p_c\bigr) \\[4pt]
  -\!\bigl(1-p_{-c}\bigr) & p_c
\end{pmatrix}.
\end{equation}

Solving for the intermediate value of $p(e|\neg\mathcal C,\mathcal S)$ and plugging this in~\cref{eq:prec_and_recall} along with~\cref{eq:correlation_correction_term}, we obtain the final equation~\cref{eq:correction_equation}:

\begin{equation}\label{eq:correction_equation_appendix}
    p(e|\mathcal S) = \underbrace{\frac{p(e|\mathcal C)\,p_{\neg\mathcal C}+p(e|\neg\mathcal C)\,(p_\mathcal C-1)}{p_\mathcal C + p_{\neg\mathcal C}-1}}_\text{independence assumption}
    +\underbrace{\delta p(e|\mathcal S) \overbrace{\left(\frac{p_\mathcal C p_{\neg\mathcal C}}{p_\mathcal C + p_{\neg\mathcal C}-1}-r_\mathcal C\right)}^{\kappa_\mathcal S} 
    +\delta p(e|\neg\mathcal S)\overbrace{\frac{(p_\mathcal C -1)p_{\neg\mathcal C}}{p_\mathcal C + p_{\neg\mathcal C}-1}}^{\kappa_{\neg\mathcal S}}}_\text{correction terms}\,.
\end{equation}

Regarding the denominator $p_\mathcal C + p_{\neg\mathcal C}-1$, it can be zero (or approximately zero) for some combinations of precision of the metadata annotation process. In these cases, no statement can be made on $\mathcal S$ as the performance of the annotation classification does not allow separation of $\mathcal S$ from the rest of the data and any observable error differences on $\mathcal C$ potentially stems only from the correction factors.
Besides this technical breakdown of the hypothesis, it should be pointed out that the scaling factors $\kappa_{\mathcal S,\neg\mathcal S}$ depend only on the performance of the annotation process and thus can be determined without knowing the correction factors $\delta p$ themselves. While the latter are challenging to determine in practice they are rarely non-zero, even in cases where the hypothesis holds, due to fluctuations (e.g.\ when errors are determined on finite sample sizes). Knowing the magnitude of $\kappa$ therefore allows us a degree of certainty on the statements of the hypothesis.

\subsection{Quantitative Evaluation of Metadata Generation Process}
\label{appendix:quantitaive_metadata_evaluation}

Our metadata generation is a form of data labeling process. 
Within this work, we chose CLIP~\citep{radford2021learning} to generate the metadata but know that for certain attributes of the ODDs the performance might be far below human capabilities, compare, e.g.,~\citet{gannamaneni2023investigating}.
To estimate the performance of our metadata generation process without large-scale evaluation or manual labeling, we take a simplifying view.
For each slice $\mathcal{C}$ containing a semantic concept identified by CLIP, for instance, images containing gender ``female'', we randomly draw a few samples to create a smaller subset $\mathcal{R}$.
Let $q$ denote the probability that images within $\mathcal{C}$ contain the correct semantic concept.
By manually evaluating the smaller sample of images $\mathcal{R}\subset\mathcal{C}$ (drawn with replacement), we can model the posterior distribution for $q$ using Bayes theorem, that is
\begin{equation}
    p(q|\mathcal{R})=\frac{p(q)p(\mathcal{R}|q)}{p(\mathcal{R})}\propto p(\mathcal{R}|q)\,.
    \label{eq:bayes}
\end{equation}
Therein, we assumed a flat prior, i.e. $p(q)=\text{const.}$.
The probability of the observed sample $\mathcal{R}$ is given by
\begin{equation}
    p(\mathcal{R}|q)=\binom{n}{l} q^l (1-q)^{n-l}\,,
    \label{eq:bernoulli}
\end{equation}
where $n=|\mathcal{R}|$ is the size of the observed sample taken from $\mathcal{S}$ and $l\leq n$ is the number of observed positive, i.e., correct instances.
The true value for $q$ for the entire slice would describe the precision of the labeling of the concept as it is the ratio of true instances to the overall number of samples. We can approximate it using the small set using
\begin{equation}
    p_\text{precision}(q|\mathcal{R}) = \frac{(n+1)!}{l!(n-l)!} q^l (1-q)^{n-l}\,,
    \label{eq:approx:precison}
\end{equation}
where the factorials serve as the normalisation.
Using~\cref{eq:approx:precison}, we can, therefore, determine both the expected value of $q$ as well as our uncertainty of its value, which we report in terms of the standard deviation $\sigma$.
As a side note, for values of $q$ near 0 or 1, the Binomial distribution is asymmetric and the standard deviation is not always a faithful measure of ``true'' deviation. However, we compared with a quantile based approach, taking the range from the $1/6^\text{th}$ to $5/6^\text{th}$ quantile, and found only minor discrepancies.

Besides estimating the precision, we are also interested in estimating the recall of the labeling process.
This latter quantity is harder to evaluate as it depends both on the number of true positives and false negatives.
Let $P$ and $N$ denote the total number of data points that are classified as containing, or respectively, as not containing, the semantic concept.
Then the probability over the total number of true positives is given by  $p_\text{precision}(q| \mathcal{R}_P)$, where $\mathcal{R}_{P}$ is a random sample taken from the set $\mathcal{C}_P$ of positively classified elements.
A similar statement holds for the number of false negatives, where a sample $\mathcal{R}_N$ from the non-detected set can be used.
However, in this case, we either have to count (for $l$) the number of prediction errors or use the inverse outcome $1-q$.
Given that both samples are free of intersection, that is $\mathcal{R}_P \cap \mathcal{R}_N = \emptyset$, we make the assumption that the obtained probabilities $q_P$ and $q_N$ are independent from one another.
In this case, we can formulate the recall as
\begin{equation}
\begin{split}
    p_\text{recall}(q|\mathcal{R}_P, \mathcal{R}_N) = & \int_0^1\!\mathrm{d}q_P\int_0^1\!\mathrm{d}q_N \\
    \times & \delta\left(q-\frac{P q_P}{P q_P + N (1-q_N)} \right) \\
    \times & p_\text{precision}(q_P|\mathcal{R}_P)p_\text{precision}(q_N|\mathcal{R}_N)\,,  
\end{split}
\label{eq:approx:recall}
\end{equation}
where we interpret $p_\text{precision}$ such that in both cases correct predictions are counted while $\delta$ denotes a Dirac-Delta Distribution.
We evaluate this function numerically and use the results of $p_\text{recall}$ in the same way as for the precision above regarding, e.g., the reported standard deviation.

\subsection{Precision Sampling at different levels}
\label{appendix:level1_precisions}

In~\cref{appendix:quantitaive_metadata_evaluation}, we provide the framework for how precision and recall can be estimated by sampling data in slices. 
In this section, we present the concrete steps taken at level 1 to operationalize it and also the steps taken to calculate precision and recall at higher levels. At level 1, in synthetic and celebA experiments, as GT labels are available in addition to classification function $\mathcal G$ labels, human evaluation of slices is not necessary. For each slice in the data, before running SliceLine, we sample with replacement (n=60), and using GT slice labels, calculate precision and recall based on~\cref{appendix:quantitaive_metadata_evaluation}. This gives us mean and standard deviations of precision and recall that can be used with~\cref{eq:correction_equation}. For AD datasets, as GT labels are not available, we performed human evaluation by first taking 60 samples for each level 1 slice. The results of this are shown in~\cref{table:estimation_metadata_quality}.

At level 2 and higher, human sampling of precisions gets very labor-intensive even if considering only 9 semantic dimensions with binary attributes. Therefore, we incorporate the parent-level precisions calculated earlier to estimate corrected errors by accounting for their contributions. From level 2 onward, we construct a composite inverse matrix,
\begin{equation}
    A^{-1}_{\mathcal{S}_1 \mathcal{S}_2}=A^{-1}_{\mathcal{S}_1} \otimes A^{-1}_{\mathcal{S}_2}\,,
    \label{eq:compositeAinv}
\end{equation}
which is a direct product of the inverse matrices given in \cref{{eq:matrix_corr_equation}} for the respective semantic dimensions $\mathcal S_1$ and $\mathcal S_2$. The direct product implies an element-wise multiplication of the differing elements of $A^{-1}_{\mathcal{S}_i}$ in all possible combinations. 
This approach can be understood by first considering that in the approximation the precision values in $A_{\mathcal{S}_1 \mathcal{S}_2}$ are given by products of the respective precisions for $\mathcal S_{1,2}$ or its negations $\neg\mathcal{S}_{1,2}$.
That is, for two $2\times 2$ matrices $A_{\mathcal{S}_i}$ the resulting $A_{\mathcal{S}_1 \mathcal{S}_2}$ will be $4\times 4$ dimensional.
Second, the inverse of this direct product matrix is given by the direct product of its constituent matrices, leading to~\cref{eq:compositeAinv}.

We can also extend the binary case of ~\cref{{eq:matrix_corr_equation}} to a multi-class setting by taking into account that the matrices $A$ are based on normalized confusion matrices. That is, row-wise the entries in $A$ give the rate or probability with which the classifier $\mathcal G$ will mistake a given element for an element of a foreign class. This notion easily generalizes to arbitrary classes by taking the full confusion matrix for all classes, thereby introducing all combinations besides ``False Positives'' or ``False Negatives'' from the binary case.
For $n$ classes this would result in a $n\times n$ matrix for $A$, the inverse of which can be used to obtain the hypothesis part of~\cref{eq:correction_equation_appendix}.

\section{Results}
\subsection{Further results: CelebA Evaluations}
\label{appendix:celeba_further_eval}
\begin{figure}[htbp]
    \centering
    \begin{minipage}{\textwidth}
      \includegraphics[width=\linewidth]{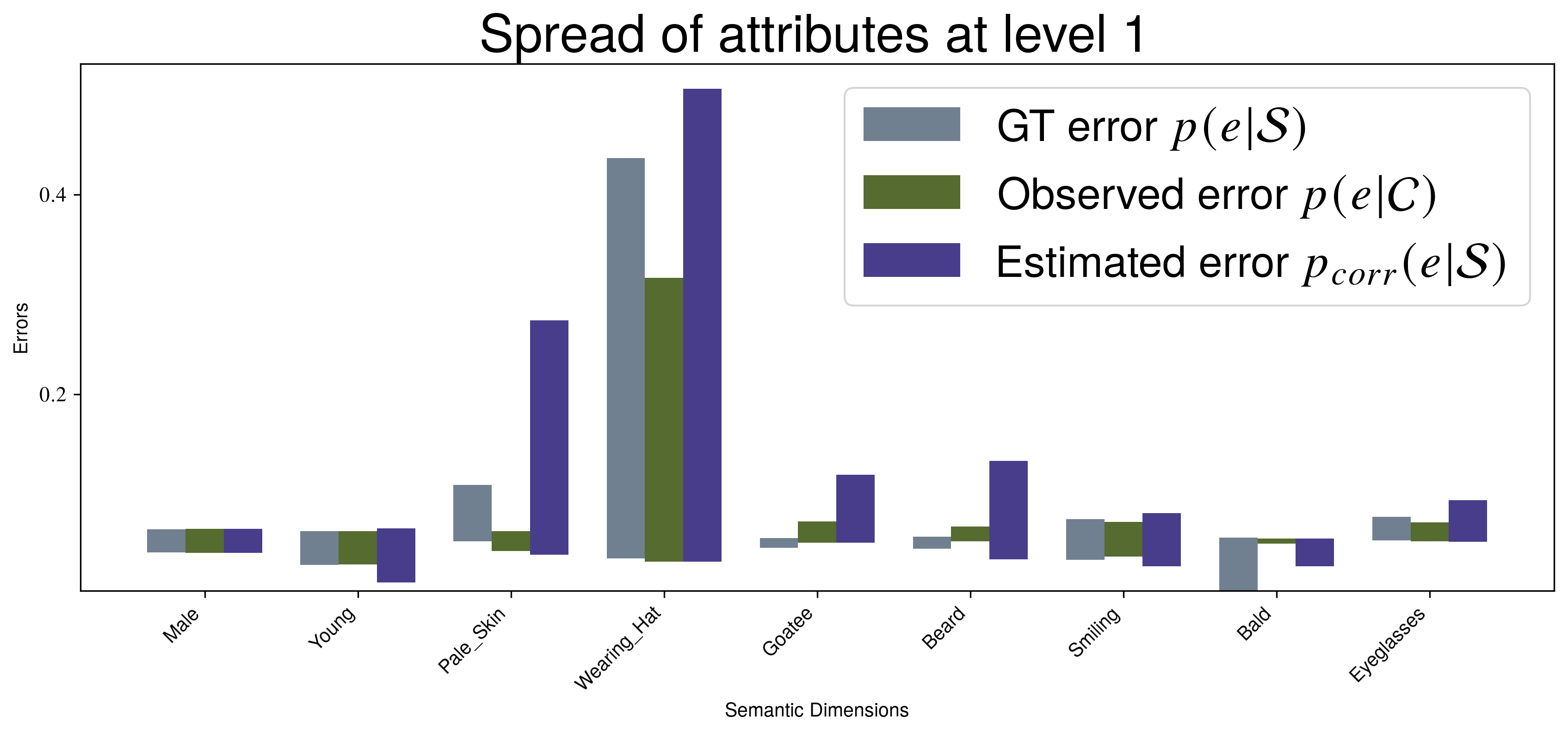} 
    \end{minipage}%
    \hfill 
    \begin{minipage}{\textwidth}
      \includegraphics[width=\linewidth]{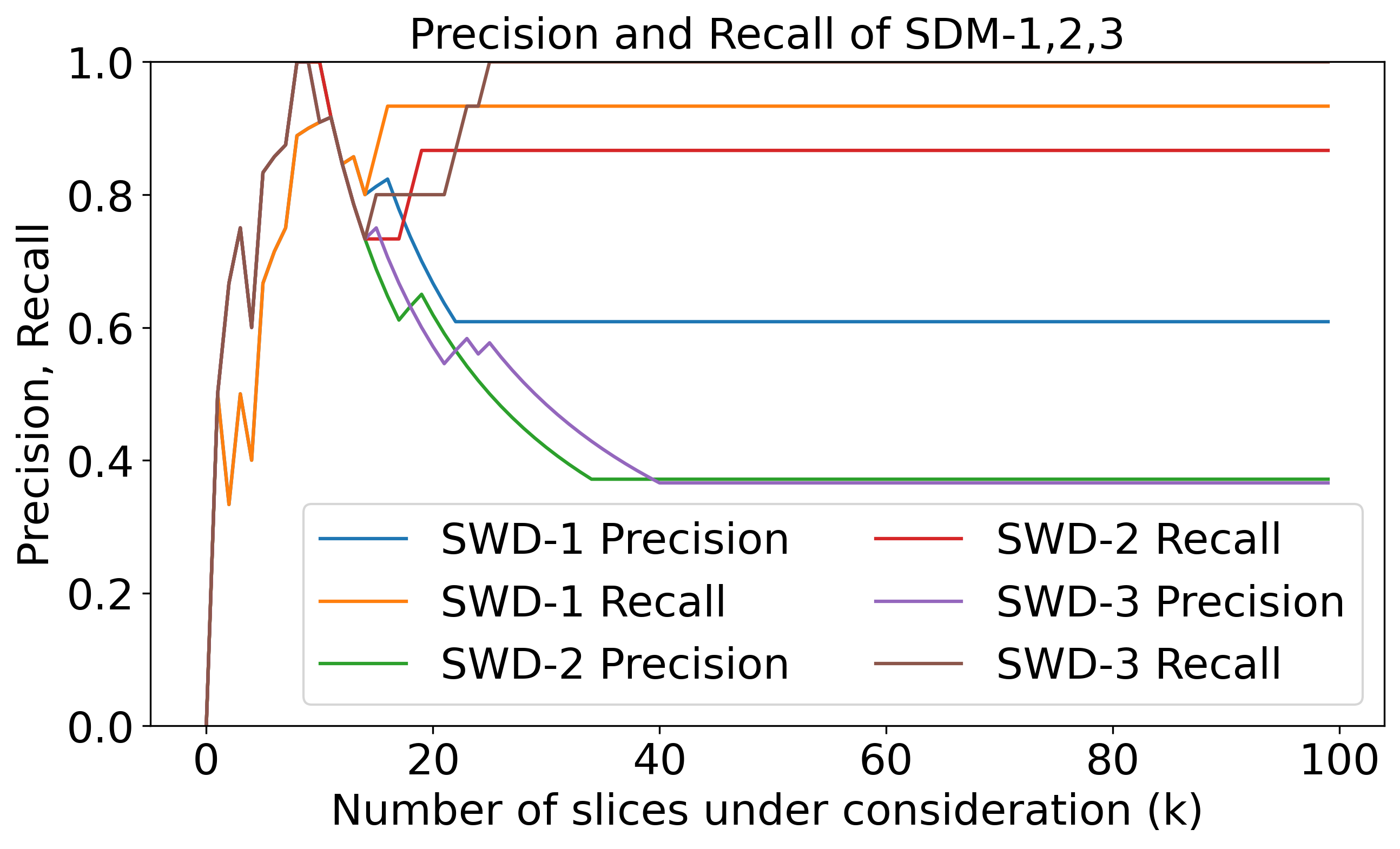} 
  \end{minipage}
    \caption{Similar to synthetic data experiment, we provide spread of error (top) and Precision and Recall at different levels of k for SWD-1,2,3 of~\cref{algo:combined_sliceline} in comparison to the Oracle (bottom). The \textbf{DuT} is a ViT-B-16 classification model trained on ImageNet21k and evaluated on celebA dataset. Note that here precision and recall are quality metrics of weak slice discovery and not of labeling quality.}
    \label{fig:celebA_furthereval}
\end{figure}

In the synthetic data experiment, we provide the spread of errors and the precision and recall of slice recovery in comparison to an Oracle for different values of $k$. With the GT metadata in the celebA dataset, we build a similar Oracle for comparison and provide similar error spread and precision and recall values of SWD-1,2,3 in~\cref{fig:celebA_furthereval}. Here, the plot depicting the spread of errors is restricted to level 1 errors for better visualization. However, the precision and recall plot is based on the full level 2 slices.

\subsection{Evaluation of Top-5 Weak Slices}
\label{appendix:qualitative_metadata_evaluation}

\begin{figure*}[htbp!]
     \centering  
     \begin{subfigure}{0.49\textwidth}
         \centering
         \includegraphics[width=\linewidth]{images/e2_s0.pdf}        
     \end{subfigure}     
     \hfill     
     \begin{subfigure}{0.49\textwidth}
         \centering
         \includegraphics[width=\linewidth]{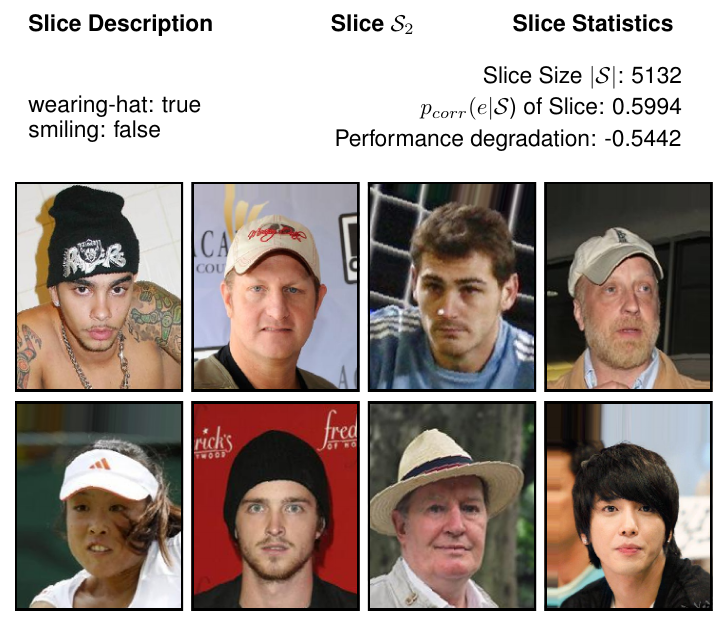}        
     \end{subfigure}
     \\
     \begin{subfigure}{0.49\textwidth}
         \centering
         \includegraphics[width=\linewidth]{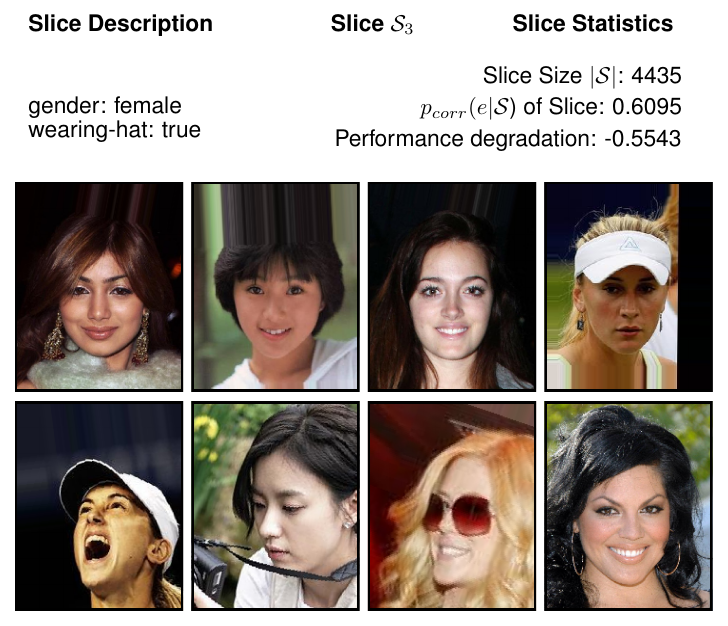}        
     \end{subfigure}     
     \hfill     
     \begin{subfigure}{0.49\textwidth}
         \centering
         \includegraphics[width=\linewidth]{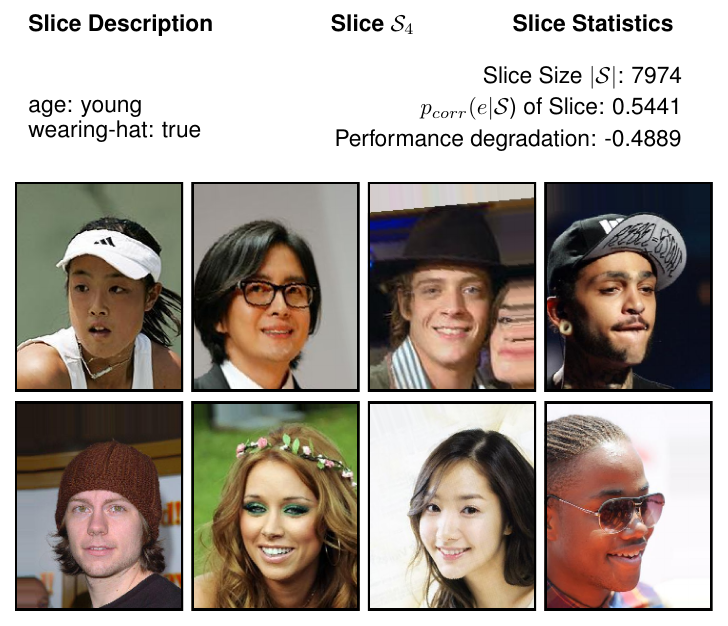}        
     \end{subfigure}
     \\
     \begin{subfigure}{0.49\textwidth}
         \centering
         \includegraphics[width=\linewidth]{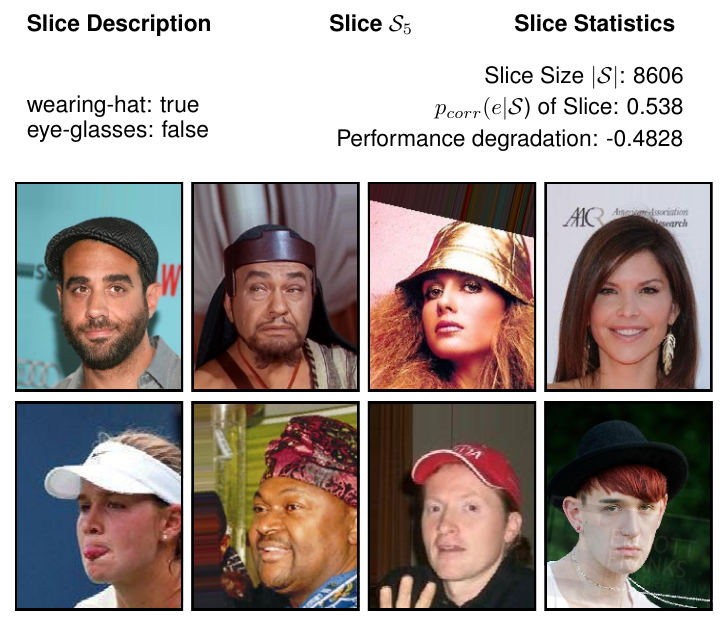}        
     \end{subfigure}     
     \caption{Samples from top-5 weak slices obtained using SWD-3 for the ViT-B-16 classification model trained on ImageNet21k and evaluated on the full celebA dataset with metadata generated from CLIP using step 3 in~\cref{algo:combined_sliceline}. The statistics provide a quantitative evaluation of the entire slice. For qualitative evaluation, we provide some sample images from the slice.}
     \label{fig:appendix_celebA_clip}
\end{figure*}

\begin{figure*}
\centering
     \includegraphics[width=0.95\textwidth]{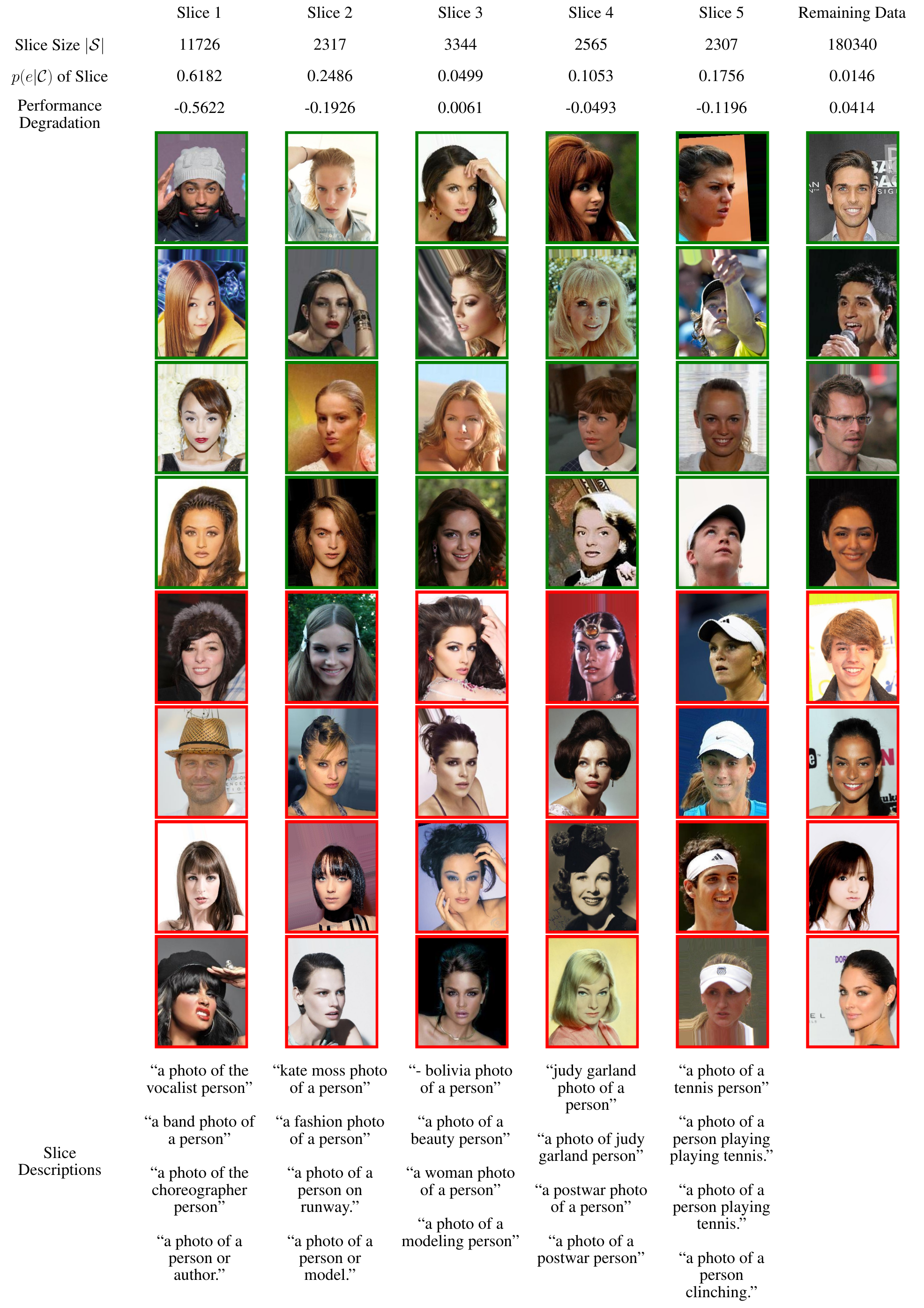}     
     \caption{Samples from top-5 weak slices of a ViT-B-16 classification model trained on ImageNet21k and evaluated on the full celebA dataset (DOMINO). From the 8 samples in each slice, 4 are true positives (green outline) and 4 are false negatives (red outline).}
     \label{fig:domino_results}     
\end{figure*}

\begin{figure*}
\centering
     \includegraphics[width=0.95\textwidth]{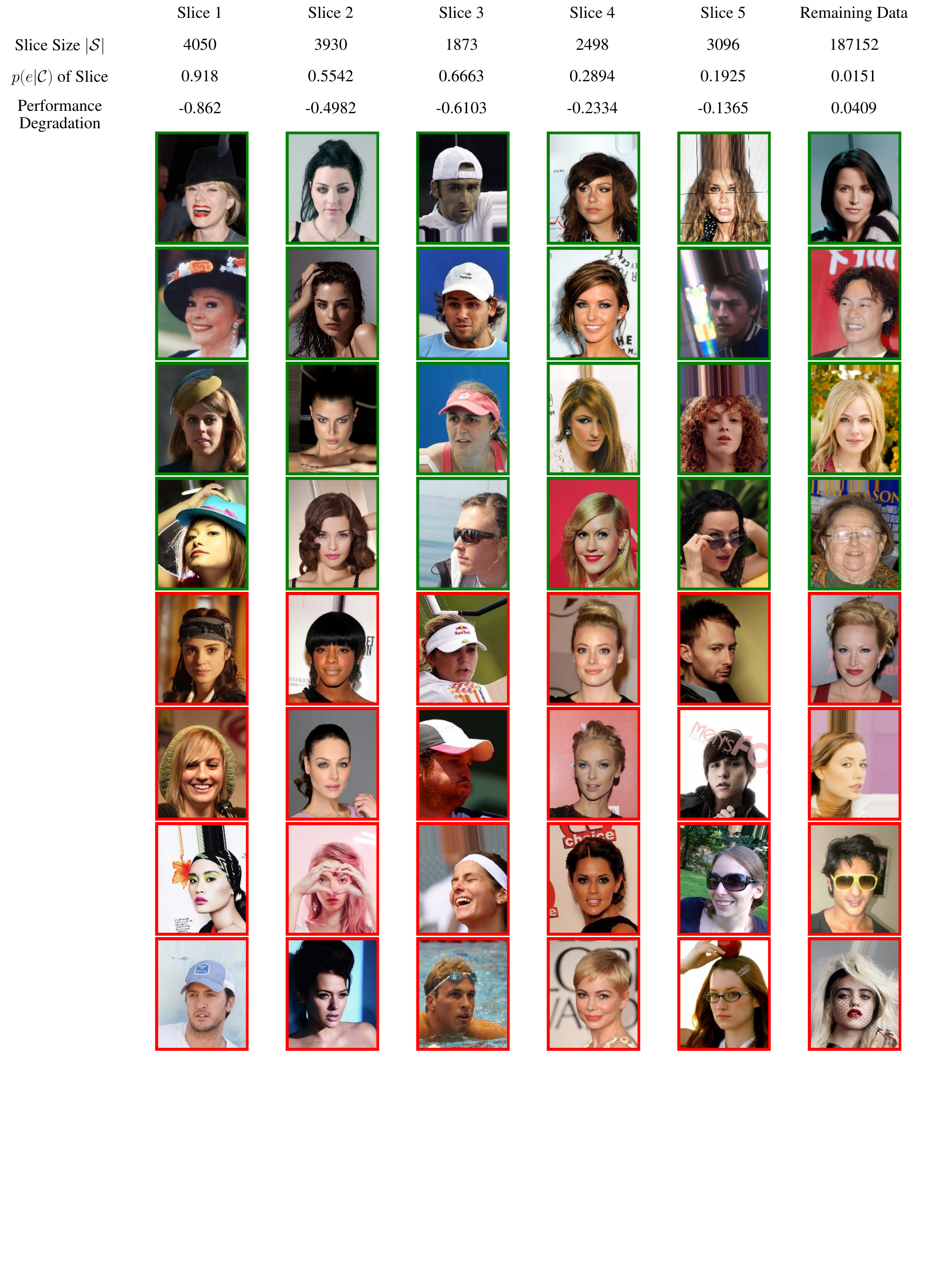}     
     \caption{Samples from top-5 weak slices of a ViT-B-16 classification model trained on ImageNet21k and evaluated on the full celebA dataset (Spotlight). From the 8 samples in each slice, 4 are true positives (green outline) and 4 are false negatives (red outline). Spotlight does not provide automatic descriptions of the slices}
     \label{fig:spotlight_results}     
\end{figure*}

\begin{figure*}
\centering
     \includegraphics[width=0.95\textwidth]{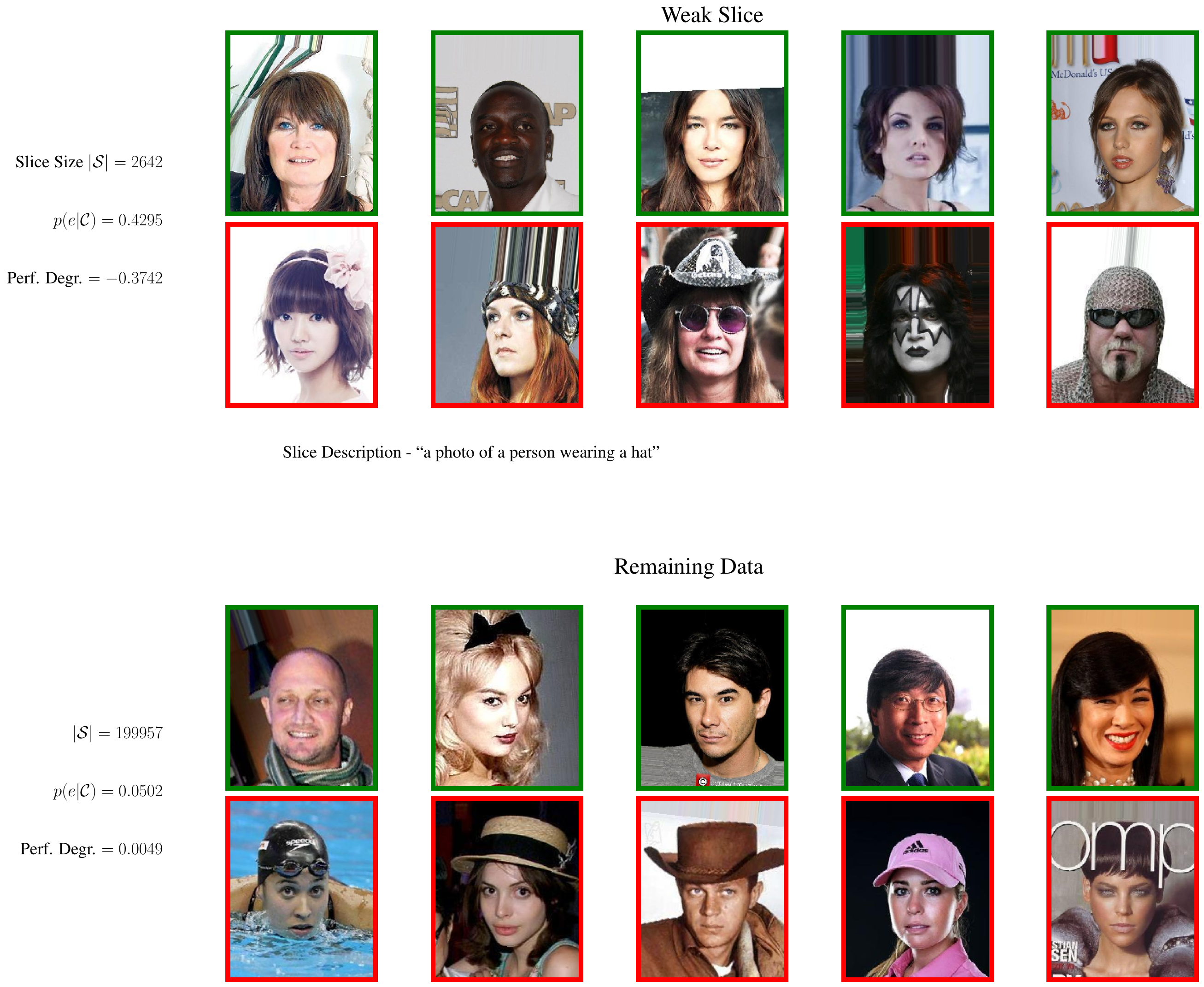}     
     \caption{Samples from top-1 weak slices of a ViT-B-16 classification model trained on ImageNet21k and evaluated on the full celebA dataset (SVM-FD). From the 10 samples in each slice, 5 are true positives (green outline) and 5 are false negatives (red outline). Unlike other SDMs, SVM-FD only outputs one weak slice.}
     \label{fig:svmfd_results}     
\end{figure*}

In this section, we provide both the quantitative and qualitative results of our experiments. For the celebA dataset, \cref{fig:appendix_celebA_clip,fig:domino_results,fig:spotlight_results,fig:svmfd_results} contain the identified top-5 weak slices in the experiments SWD-3, DOMINO, Spotlight, and SVM FD respectively. We provide 8 samples from each of the top-5 slices found by the methods and 8 samples from the remaining data, except SVM FD which only provides 1 weak slice. In addition, we provide four slice descriptions given by DOMINO for each slice and the single slice description of SVM FD. While the actionability of our proposed approach is inherent as the identified weak slices are based on semantic concepts from the ODD, the textual descriptions from DOMINO are comparatively less useful. Furthermore, by focusing only on the samples from DOMINO, it is still hard to identify which semantic concepts uniquely constitute a slice. For example, if we consider an image from the remaining data (rightmost column), it is not straightforward to say if this image does or does not belong to any of the weak slices. Although the fifth slice does appear to capture a coherent slice, images of sports persons, the observed error $p(e|\mathcal{C})$ is significantly lower than what is identified by our approach. 
It is important to note that both DOMINO and SliceLine judge performance in terms of class probabilities, not false negative counts. Therefore, weak slices can have slightly better performance in terms of $p(e|\mathcal{C})$ compared to overall data, as observed for slice 3 found by DOMINO.

\begin{figure*}[htbp!]
     \centering  
     \begin{subfigure}{0.49\textwidth}
         \centering
         \includegraphics[width=\linewidth]{images/e3_s0.pdf}        
     \end{subfigure}     
     \hfill     
     \begin{subfigure}{0.49\textwidth}
         \centering
         \includegraphics[width=\linewidth]{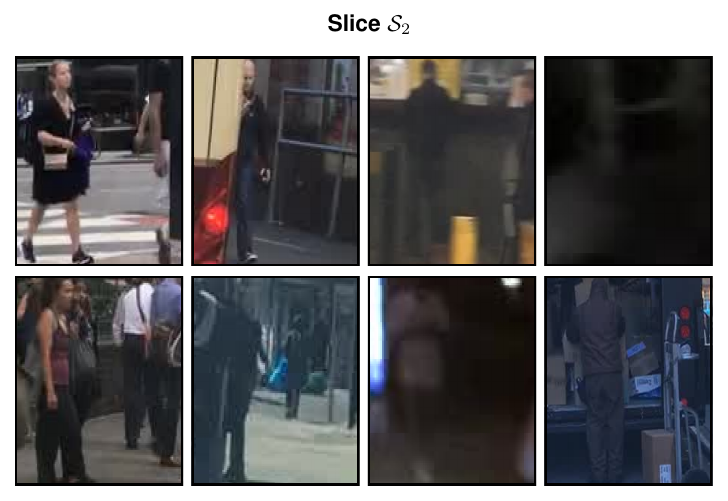}        
     \end{subfigure}
     \\
     \begin{subfigure}{0.49\textwidth}
         \centering
         \includegraphics[width=\linewidth]{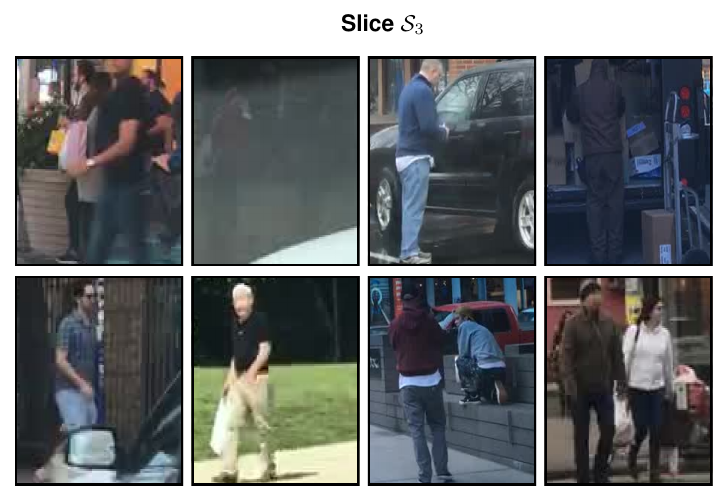}        
     \end{subfigure}     
     \hfill     
     \begin{subfigure}{0.49\textwidth}
         \centering
         \includegraphics[width=\linewidth]{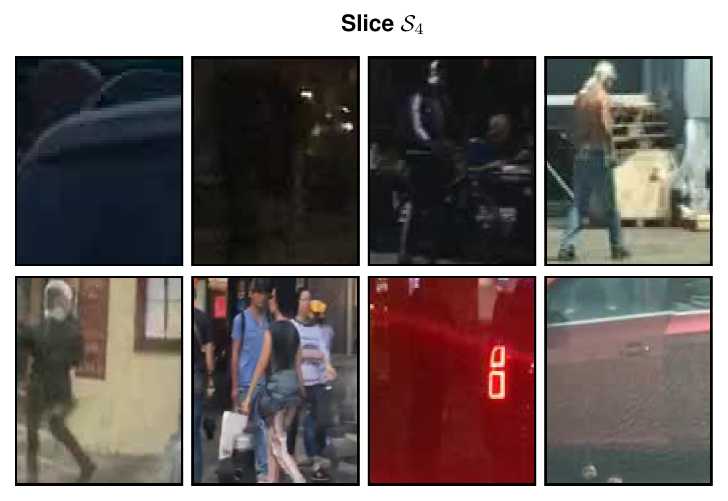}        
     \end{subfigure}
     \\
     \begin{subfigure}{0.49\textwidth}
         \centering
         \includegraphics[width=\linewidth]{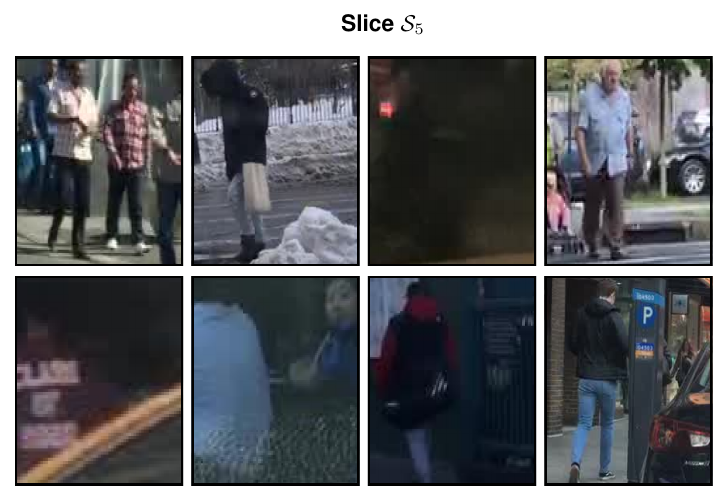}        
     \end{subfigure}     
     \caption{Samples from top-5 weak slices obtained using SWD-3 for the Faster R-CNN object detector trained and evaluated on BDD100k dataset.}
     \label{fig:appendix_bdd100k}
\end{figure*}

\begin{figure*}[htbp!]
     \centering  
     \begin{subfigure}{0.49\textwidth}
         \centering
         \includegraphics[width=\linewidth]{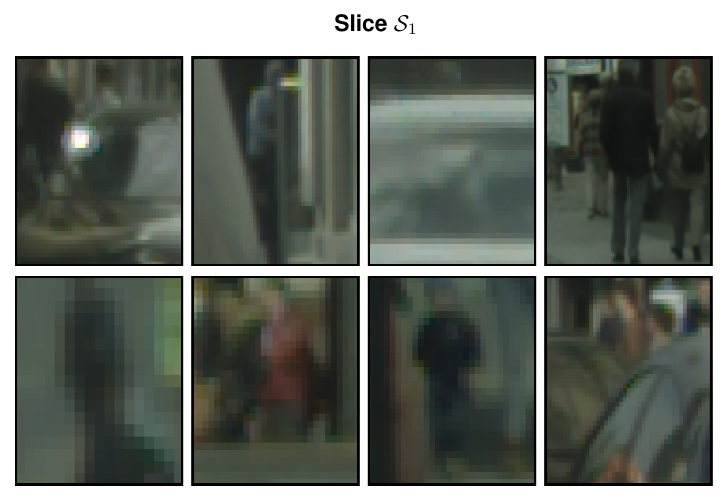}        
     \end{subfigure}     
     \begin{subfigure}{0.49\textwidth}
         \centering
         \includegraphics[width=\linewidth]{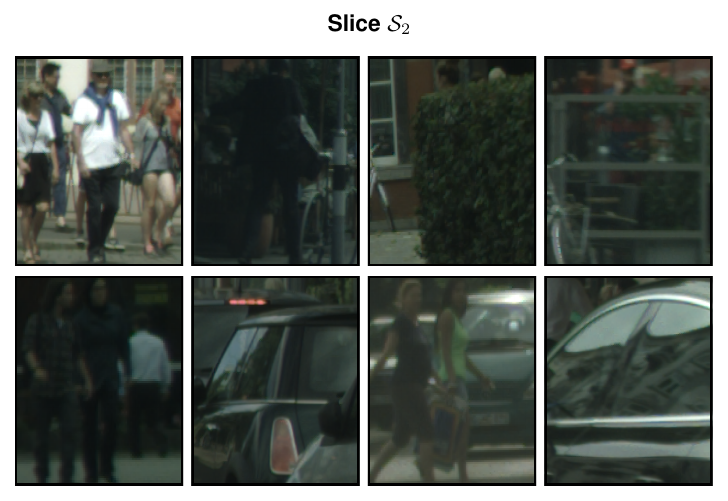}        
     \end{subfigure}
     \\
     \begin{subfigure}{0.49\textwidth}
         \centering
         \includegraphics[width=\linewidth]{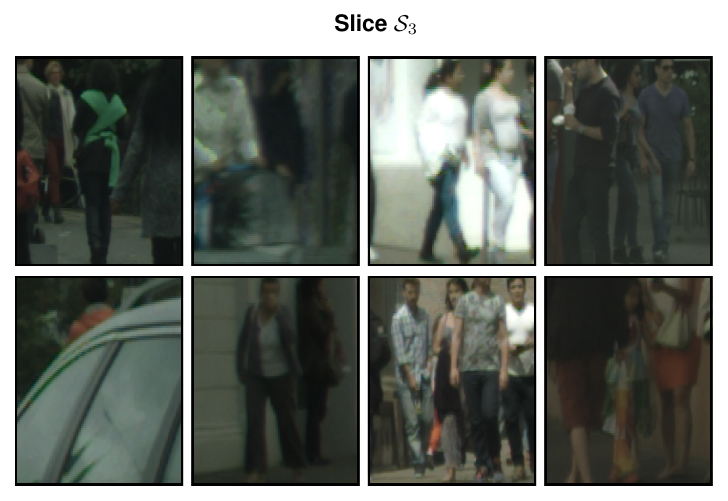}        
     \end{subfigure}     
     \begin{subfigure}{0.49\textwidth}
         \centering
         \includegraphics[width=\linewidth]{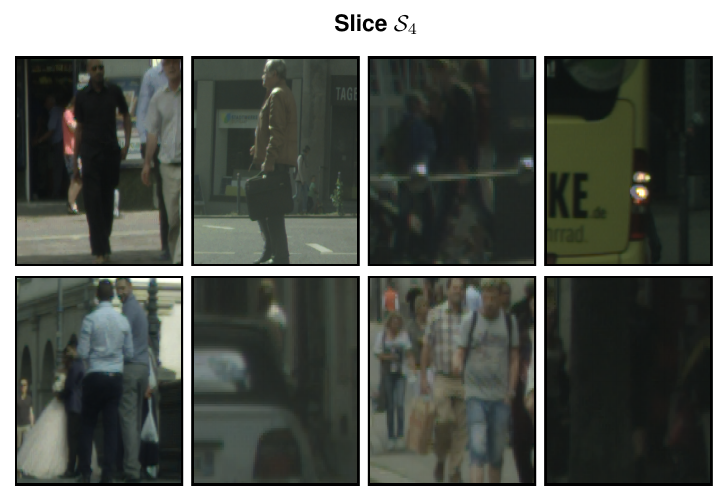}        
     \end{subfigure}
     \\
     \begin{subfigure}{0.49\textwidth}
         \centering
         \includegraphics[width=\linewidth]{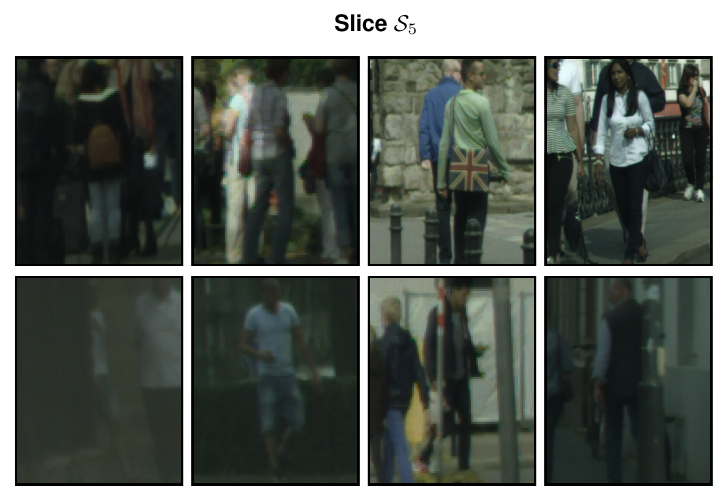}        
     \end{subfigure}

     \caption{Samples from top-5 weak slices obtained using SWD-3 for the SETR semantic segmentation model trained and evaluated on Cityscapes dataset.}
     \label{fig:appendix_cityscapes}
\end{figure*}

\begin{figure*}[htbp!]

     \centering  
     \begin{subfigure}{0.49\textwidth}
         \centering
         \includegraphics[width=\linewidth]{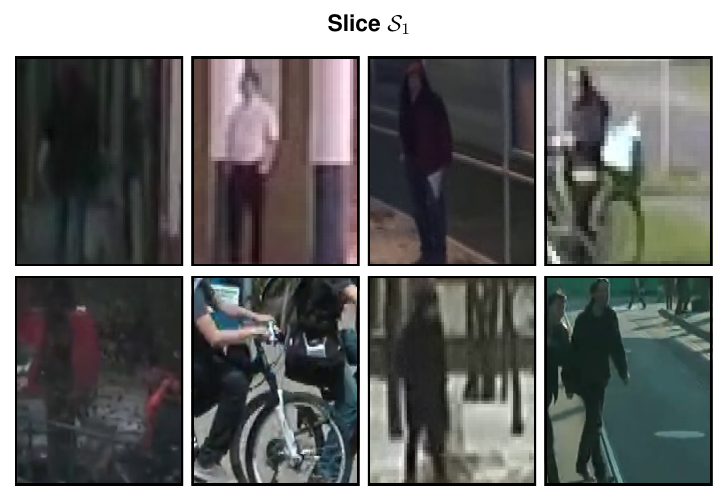}        
     \end{subfigure}     
     \hfill     
     \begin{subfigure}{0.49\textwidth}
         \centering
         \includegraphics[width=\linewidth]{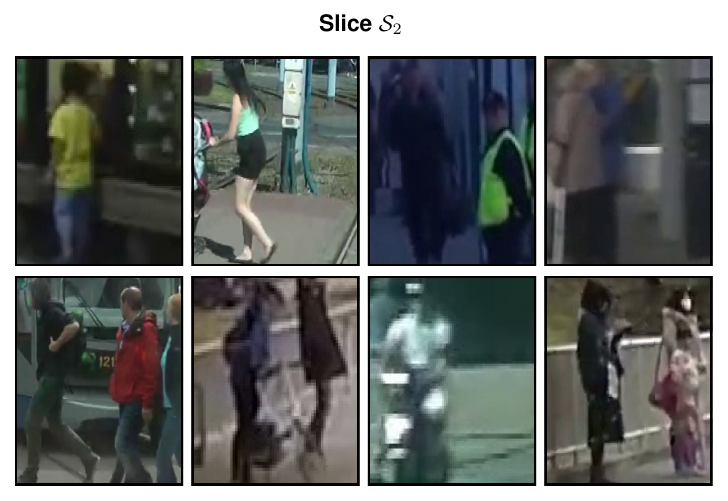}        
     \end{subfigure}
     \\
     \begin{subfigure}{0.49\textwidth}
         \centering
         \includegraphics[width=\linewidth]{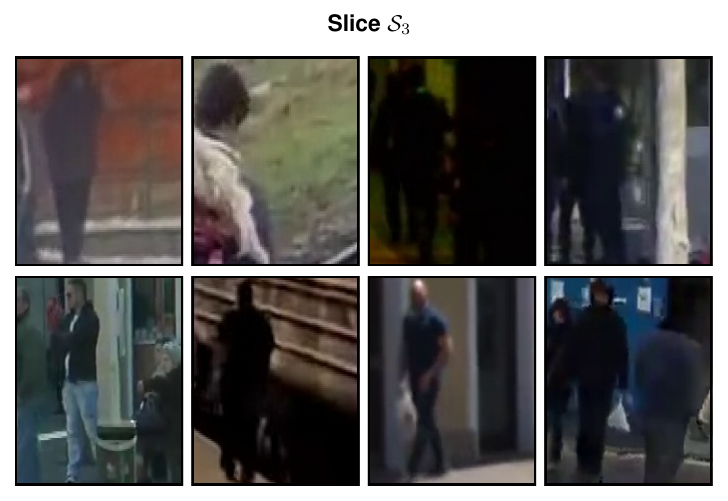}        
     \end{subfigure}     
     \hfill     
     \begin{subfigure}{0.49\textwidth}
         \centering
         \includegraphics[width=\linewidth]{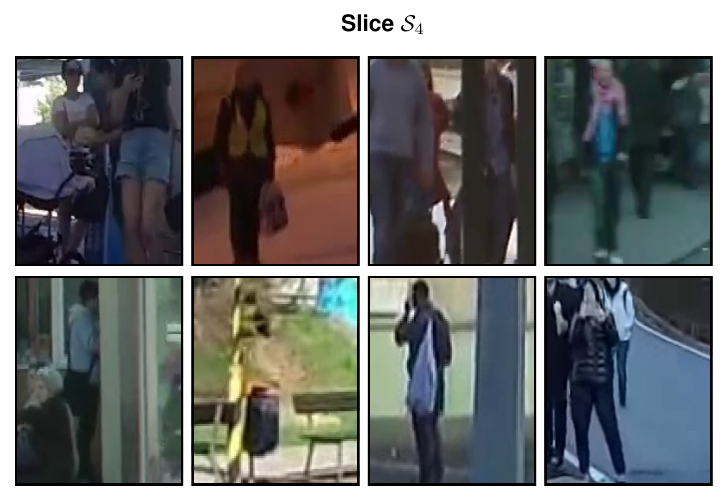}        
     \end{subfigure}
     \\
     \begin{subfigure}{0.49\textwidth}
         \centering
         \includegraphics[width=\linewidth]{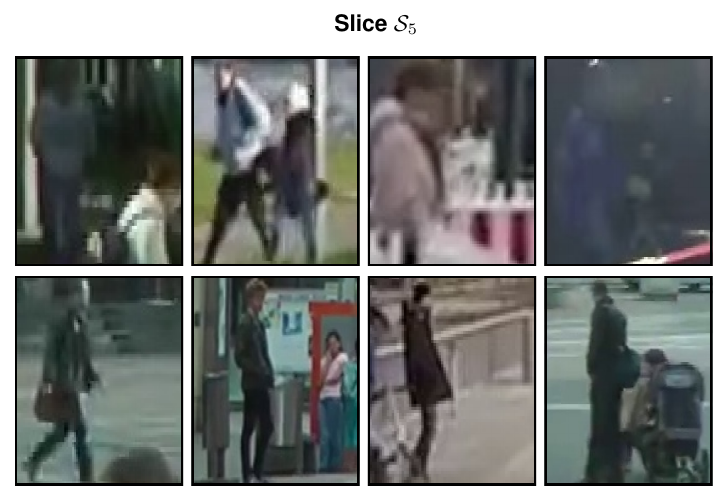}        
     \end{subfigure}         
     
     \caption{Samples from top-5 weak slices obtained using SWD-3 for the Panoptic-FCN model trained and evaluated on RailSem19 dataset.}
     \label{fig:appendix_railsem}
\end{figure*}

In~\cref{fig:appendix_bdd100k,fig:appendix_cityscapes,fig:appendix_railsem}, pedestrian crop samples from the top-5 weak slices obtained using our method are provided for each autonomous driving experiment. The quantitative evaluation of the top-5 slices for the three experiments can be found in~\cref{tab:appendix:bdd100k,tab:appendix:cityscapes,tab:appendix:railsem}. 


\begin{table}[htbp!]
\setlength{\tabcolsep}{3pt}

    \centering
    \begin{tabular}{c|cccl}
    Slice No. & $\mathcal{|S|}$ & $p_\text{corr}(e|\mathcal{S})$ & \makecell{Avg. Perf. \\ Degra.} & Slice Description \\
    \hline
    $\mathcal{S}_1$   &  319	& 0.2206   & -0.1636 & \messagebubble{\makecell[l]{\textbf{blurry}: false \\
\textbf{occluded}: true}} \\
\hline
    $\mathcal{S}_2$   &  508	& 0.2099   & -0.1528 & \messagebubble{\makecell[l]{\textbf{blurry}: false \\
    \textbf{cloth.-color}: dark-color \\
}} \\
\hline

    $\mathcal{S}_3$   &  466	& 0.147	   & -0.0899 & \messagebubble{\makecell[l]{\textbf{blurry}: false \\
\textbf{age}: adult \\
}} \\
\hline

    $\mathcal{S}_4$   &  773	& 0.1263   & -0.0693 & \messagebubble{\makecell[l]{\textbf{blurry}: false}} \\
\hline

    $\mathcal{S}_5$   &  582	& 0.1263   & -0.0693 & \messagebubble{\makecell[l]{\textbf{blurry}: false \\
    \textbf{gender}: Male
}} \\

    \end{tabular}
    \caption{Quantitative analysis of the top-5 weak slices obtained using SWD-3 for the Faster R-CNN object detector trained and evaluated on BDD100k dataset.}
    \label{tab:appendix:bdd100k}
\end{table}

\begin{table}[htbp!]
\setlength{\tabcolsep}{3pt}

    \centering
    \begin{tabular}{c|cccl}
    Slice No. & $\mathcal{|S|}$ & $p_\text{corr}(e|\mathcal{S})$ & \makecell{Avg. Perf. \\ Degra.} & Slice Description \\
        \hline

    $\mathcal{S}_1$     & 690 & 0.1046 & -0.0897 & \messagebubble{\makecell[l]{\textbf{age}: adult \\
        \textbf{skin-color}: dark}} \\       
    \hline

    $\mathcal{S}_2$     & 591 & 0.0921 & -0.0773 & \messagebubble{\makecell[l]
    {
        \textbf{skin-color}: dark \\
        \textbf{cloth.-color}: dark-color}} \\
    \hline

    $\mathcal{S}_3$     & 349 & 0.0896 & -0.0748 & \messagebubble{\makecell[l]{\textbf{gender}: female \\
        \textbf{skin-color}: dark \\
        }}\\
    \hline

    $\mathcal{S}_4$     & 766 & 0.0778 & -0.0630 & \messagebubble{\makecell[l]
    {\textbf{skin-color}: dark \\
       \textbf{blurry}: false}}\\
    \hline

    $\mathcal{S}_5$     & 997 & 0.0594 & -0.0446 & \messagebubble{\makecell[l]{\textbf{skin-color}: dark \\
        }} \\
    \end{tabular}
    \caption{Quantitative analysis of the top-5 weak slices obtained using SWD-3 for the SETR semantic segmentation model trained and evaluated on Cityscapes dataset.}
    \label{tab:appendix:cityscapes}
\end{table}

\begin{table}[htbp!]
\setlength{\tabcolsep}{3pt}

    \centering
    \begin{tabular}{c|cccl}
    Slice No. & $\mathcal{|S|}$ & $p_\text{corr}(e|\mathcal{S})$ & \makecell{Avg. Perf. \\ Degra.} & Slice Description \\
        \hline

     $\mathcal{S}_1$     & 541 & 0.8663 & -0.222 & \messagebubble{\makecell[l]{\textbf{age}: young}} \\      
    \hline

    $\mathcal{S}_2$     & 510 & 0.8723 & -0.228 & \messagebubble{\makecell[l]{\textbf{age}: young \\
        \textbf{construction-worker}: false}} \\       
    \hline

    $\mathcal{S}_3$     & 405 & 0.8819 & -0.2376 & \messagebubble{\makecell[l]{\textbf{skin-color}: dark \\
        \textbf{cloth.-color}: dark-color}} \\       
    \hline

    $\mathcal{S}_4$     & 349 & 0.9095 & -0.2652 & \messagebubble{\makecell[l]{\textbf{age}: young \\
        \textbf{blurry}: false}} \\       
    \hline

    $\mathcal{S}_5$     & 173 & 1.00 & -0.4602 & \messagebubble{\makecell[l]{\textbf{age}: young \\
        \textbf{skin-color}: dark}} \\       
    \hline

    \end{tabular}
    \caption{Quantitative analysis of the top-5 weak slices obtained using SWD-3 for the Panoptic-FCN model trained and evaluated on RailSem19 dataset.}
    \label{tab:appendix:railsem}
\end{table}

\end{document}